\documentclass[conference]{IEEEtran}
\usepackage{times}

\usepackage{multicol}
\usepackage[bookmarks=true]{hyperref}
\usepackage[utf8]{inputenc}
\usepackage{xcolor}
\usepackage[numbers,sort]{natbib}
\usepackage{graphicx}
\usepackage{amsmath,amssymb}
\usepackage{textcomp}
\usepackage{caption}
\usepackage{booktabs}
\definecolor{MyDarkGreen}{rgb}{0.0, 0.5, 0.0} 
\newcommand{\mypara}[1]{\par\vspace*{0mm} \textbf{\underline{{#1}}}}
\newcommand{\Revise}[1]{{#1}}

\begin{document}

% paper title
\title{FERMI: Flexible Radio Mapping with \\ a Hybrid Propagation Model and \\ Scalable Autonomous Data Collection}

% You will get a Paper-ID when submitting a pdf file to the conference system
% \author{Author Names Omitted for Anonymous Review. Paper-ID [67]}

\author{\authorblockN{Yiming Luo$^{1, 2}$, Yunfei Wang$^{3}$, Hongming Chen$^{4}$, Chengkai Wu$^{3}$, Ximin Lyu$^{4}$,\\ Jinni Zhou$^{3}$, Jun Ma$^{3}$, Fu Zhang$^{2}$, Boyu Zhou$^{1\dag}$} 
${}^{1}$Southern University of Science and Technology, ${}^{2}$ The University of Hong Kong, \\ ${}^{3}$Hong Kong University of Science and Technology (Guangzhou), ${}^{4}$ Sun Yat-Sen University}
% avoiding spaces at the end of the author lines is not a problem with
% conference papers because we don't use \thanks or \IEEEmembership

% \maketitle

\IEEEpeerreviewmaketitle
\twocolumn[{%
	\renewcommand\twocolumn[1][]{#1}%
	\maketitle
	\begin{center}
	\end{center}
        \vspace{-10mm}
}]

\footnotetext{\dag Corresponding author: Boyu Zhou, zhouby@sustech.edu.cn.}

\begin{abstract}

Communication is fundamental for multi-robot collaboration, with accurate radio mapping playing a crucial role in predicting signal strength between robots. However, modeling radio signal propagation in large and occluded environments is challenging due to complex interactions between signals and obstacles. Existing methods face two key limitations: they struggle to predict signal strength for transmitter-receiver pairs not present in the training set, while also requiring extensive manual data collection for modeling, making them impractical for large, obstacle-rich scenarios. To overcome these limitations, we propose FERMI, a flexible radio mapping framework. FERMI combines physics-based modeling of direct signal paths with a neural network to capture environmental interactions with radio signals. This hybrid model learns radio signal propagation more efficiently, requiring only sparse training data. Additionally, FERMI introduces a scalable planning method for autonomous data collection using a multi-robot team. By increasing parallelism in data collection and minimizing robot travel costs between regions, overall data collection efficiency is significantly improved. Experiments in both simulation and real-world scenarios demonstrate that FERMI enables accurate signal prediction and generalizes well to unseen positions in complex environments. It also supports fully autonomous data collection and scales to different team sizes, offering a flexible solution for creating radio maps. Our code is open-sourced at \href{https://github.com/ymLuo1214/Flexible-Radio-Mapping}{https://github.com/ymLuo1214/Flexible-Radio-Mapping}.

\end{abstract}

\begin{figure}[t]
    \centering
    \includegraphics[width=0.98\linewidth, trim=2mm 2mm 2mm 2mm, clip]{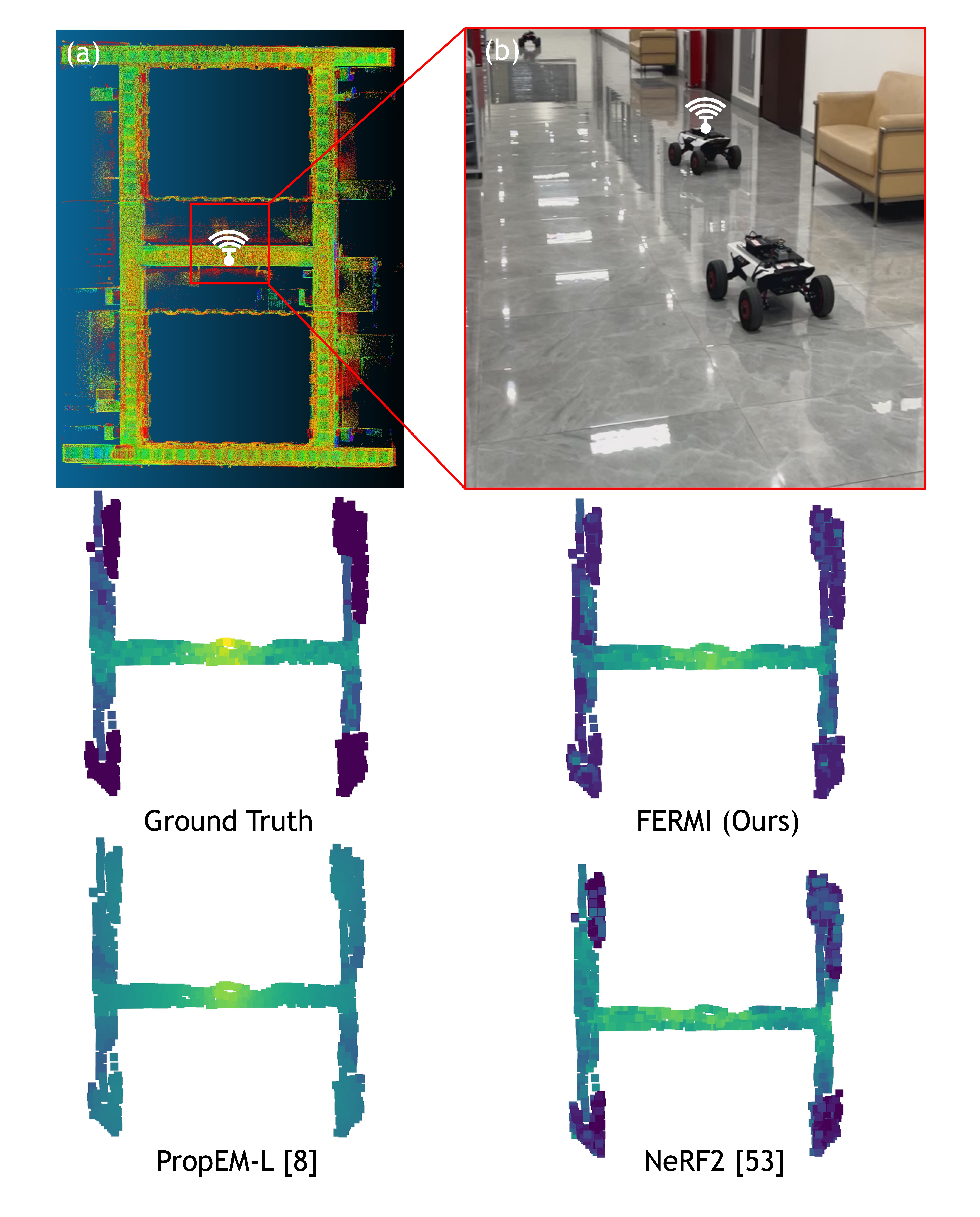}
    \caption{\textbf{Flexible Radio Mapping (FERMI)} is a data-driven, accurate, and efficient framework for signal data collection and propagation modeling. The framework enables the prediction of signal strength between any two positions using only sparse data. We demonstrate radio mapping results in an indoor corridor environment (94m × 64m × 3m), where (a) illustrates the point cloud map of the scene, (b) shows a photograph of the environment alongside our robotic platform, and below compares signal data generated by different methods when a Wi-Fi transmitter is placed at the marked position. Ground truth data is collected through dense coverage by the robots.} 
    \vspace{-5mm}
    \label{fig:teaser}
\end{figure}

\section{Introduction}

Communication plays a critical role in multi-robot cooperative tasks such as exploration and environment inspection \cite{kulkarni2022autonomous, yan2013survey}. Stable communication is essential not only for information sharing and task allocation within robot teams \cite{cao2023representation, zhou2023racer} but also for effective interaction between robots and human operators \cite{tian2024ihero}. This demand has driven the development of adaptive communication strategies using robot routers \cite{fink2013robust, yan2012robotic}. However, the effectiveness of such strategies is significantly constrained by the current limitations in radio mapping.

Radio mapping, which aims to predict signal strength at a receiver based on the positions of the transmitter (Tx) and receiver (Rx), relies on accurately modeling wireless signal propagation in the environment. In occlusion-rich and large-scale scenarios, such as subterranean environments in the DARPA SubT Challenge or complex indoor scenes, wireless signals often experience severe attenuation and complex interactions with obstacles (e.g. reflection, diffraction). These factors result in a highly intricate and discontinuous signal strength distribution, making accurate modeling significantly challenging. Moreover, as wireless signals interact with various environmental surfaces, extensive data is required for calibration to account for the complex and diverse surface properties. These challenges make existing radio mapping solutions limited to scenarios with simple obstructions.

To build a practical radio mapping framework for complex real-world environments, the following two critical characteristics are essential:
\begin{itemize}

\item\textbf{Accurate and flexible modeling:} With a reasonable amount of calibration data, the radio map can precisely determine signal strength for any given signal transmitter-receiver pair within the scene.

\item\textbf{Autonomous construction}: To achieve efficient radio mapping, a team of mobile robots should autonomously gather the necessary data in a coordinated and efficient manner, eliminating the need for labor-intensive manual deployment of transmitters (Txs) and receivers (Rxs).

\end{itemize}

However, previous methods still face significant challenges in these two areas. In the following, we identify and analyze critical issues. 

Current radio mapping approaches struggle to balance accuracy and data efficiency, largely due to the challenges inherent in modeling signal propagation. Specifically, signal propagation modeling methods can be broadly categorized into physical models \cite{bahl2000radar, kotaru2015spotfi} and data-driven methods \cite{hahnel2006gaussian, zhao2023nerf2}. Physical models capture environment-signal interactions with minimal data, enabling flexible predictions for any transmitter-receiver (Tx-Rx) pairs. However, they sacrifice accuracy due to oversimplifications, such as relying primarily on distance-based attenuation and neglecting more complex effects. In contrast, data-driven methods achieve higher accuracy by learning signal distributions from collected data. However, they require extensive data to cover the vast combinations of Tx-Rx pairs in a continuous space for training. When faced with unseen Tx-Rx pairs, these models struggle to generalize.

For autonomous data collection, it suffers from a lack of efficient and scalable planning strategies. Some approaches adopt a leader-follower paradigm, where a leader robot fits a prior model for signal data online to guide follower robots to nearby locations for data collection \cite{fink2010online, banfi2017multirobot}. However, this strategy relies on a single robot acting as the central planner, which restricts the team's ability to disperse fully throughout the environment. Additionally, fitting the prior model requires extra computational time, further limiting the overall efficiency of the task. Other methods, such as those proposed in \cite{hsieh2004constructing}, frame this task as an optimal coverage problem using a collection graph. While the problem formulation is well-defined, these methods are not scalable, offering solutions only for two robots and relying on a greedy strategy for three, which cannot be effectively extended to larger robot teams.

\begin{figure*}[t] 
    \centering
    \includegraphics[width=\linewidth,trim=1mm 1mm 1mm 1mm, clip]{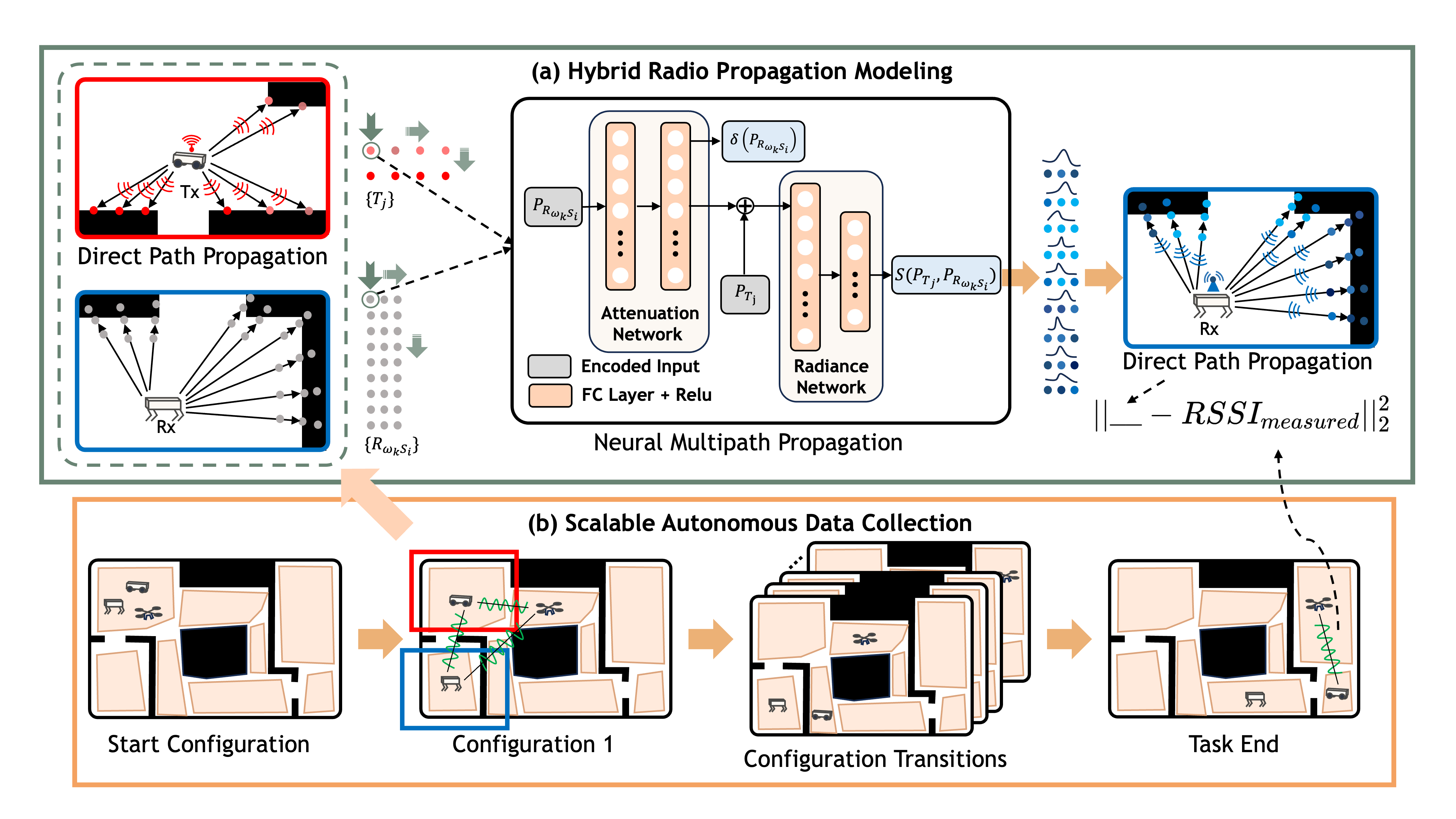}
    \caption{Overview of FERMI: (a) With the collected signal data, we first sample Line-of-Sight (LOS) points for both the transmitter (Tx) and receiver (Rx). Based on a physical model, we compute the direct path propagation from the Tx to its surrounding LOS points. Next, a neural network is employed to learn the signal strength generated by multipath propagation between each pair of LOS points from the Tx to the Rx, as well as the attenuation parameters of direct path propagation from the Rx LOS points to the Rx itself. Finally, the complex-valued signals from the Rx LOS points to the Rx are aggregated. The resulting values are compared with the signal strength data collected by the robots to compute the loss, which is used to train the multipath propagation network. (b) The autonomous data collection begins by dividing the scene into multiple regions that are mutually non-visible. A team of robots equipped with Wi-Fi access nodes departs from their initial positions, collaboratively planning to execute the most parallel and cost-effective transition sequences to traverse different regions and collect signal data.}
    \vspace{-5mm}
    \label{fig:pip}
\end{figure*}

To address the aforementioned challenges, we propose flexible radio mapping (FERMI), a framework for autonomous radio mapping using a multi-robot system. The key contributions of our work are as follows:

\textbf{Hybrid radio propagation modeling:}
 To precisely determine the signal strength for arbitrary Tx-Rx pair, we propose a hybrid approach to model signal propagation within complex environments. The propagation process is divided into three stages: direct path propagation from the Tx to its Line-of-Sight (LOS) points, multipath propagation among the LOS points of the Tx and Rx within the environment, and direct path propagation from the Rx LOS points to the Rx. The first and last stages are explicitly computed using physics-based models, where we identify LOS points for the Tx and Rx and compute their signal propagation efficiently using known physics-based attenuation. The intermediate stage, which involves complex multipath propagation, is modeled using a neural network. The network is designed to be agnostic to the specific positions of the Tx and Rx, enabling it to generalize and predict signal strength contributions from multipath effects for Tx-Rx pairs not included in the training data. We also introduce a carefully designed training architecture to learn effectively from sparse data.

\textbf{Scalable autonomous data collection:}
We further propose a scalable autonomous data collection strategy utilizing multiple robots to efficiently gather  adequate data for training the radio map. The core idea involves partitioning the scene into mutually invisible regions, where radio signals are expected to experience multipath propagation across different regions. Given the scene partitioning, robots are deployed to ensure that any pair of regions is occupied and traversed by two robots, functioning as the transmitter (Tx) and receiver (Rx) simultaneously for data collection. A formal problem formulation is provided, along with a corresponding solution that maximizes parallelism in data collection while minimizing transition costs between team configurations. This approach ensures both effective and efficient data collection.
        
Our proposed framework provides an efficient and flexible solution to enhance collaborative multi-robot systems, enabling autonomous radio mapping without human intervention. We validate the framework through both simulation and real-world experiments. In environments with significant obstructions and complexity, the results demonstrate that sampling at a sparse resolution (approximately 1–2 m grid spacing) achieves low prediction errors, reducing errors by up to 40\% compared to benchmark methods. Furthermore, we deploy a three-robot team in a real-world scenario for autonomous data collection. The framework completes data collection in a large-scale environment within 1 hour, achieving a mean prediction error of within 7 
dBm.

\section{Related Works}

\subsection{Radio Propagation Modeling}
Radio map construction has drawn increased attention in both robotics and wireless communication communities. Many multi-robot systems rely on radio signal models to maintain communication and exchange information \cite{fink2013robust, stump2008connectivity, tardioli2010enforcing, tekdas2010robotic, yang2023minimally}. Despite this, these approaches often simplify the communication model into deterministic communication radius models \cite{dixon2009maintaining, sabattini2013decentralized, guo2018multirobot} or line-of-sight (LOS) models \cite{banfi2018optimal, campos2023navigation, stump2011visibility, tekdas2010robotic, xia2023relink}. Such simplifications fail to capture interactions between signal waves and the environment, including attenuation and multipath effects, resulting in overly optimistic or conservative predictions in obstructed environments. To account for environmental influences on signals, some methods model signal propagation loss and estimate parameters to capture attenuation caused by obstacles \cite{bahl2000radar, shin2014mri, yan2012robotic}. Nevertheless, these methods exhibit high errors in obstacle-rich environments due to their neglect of multipath effects and the diverse attenuation characteristics of obstacles with varying material properties. Alternatively, ray tracing approaches simulate and estimate multipath propagation, achieving realistic simulation results \cite{hoydis2023sionna, hoydis2024learning, ruah2024calibrating}. However, these techniques require extensive material parameter modeling and are computationally demanding. 

In contrast, data-driven methods provide a promising alternative by leveraging collected signal data to model signal strength distribution or propagation processes. Gaussian processes, for instance, are frequently employed to fit and predict signal field distributions \cite{fink2010online, schwaighofer2003gpps, hahnel2006gaussian, miyagusuku2016improving, yang2024integrating}. Yet, these methods struggle with highly discontinuous signal distributions caused by dense obstacles and face significant challenges when scaling to large datasets. Neural networks, by comparison, excel at fitting complex signal distributions and handling large datasets \cite{orekondy2023winert, lu2024deep}. Some studies use neural networks as encoders or decoders to model signal distributions \cite{lee2024scalable, Clark-RSS-22, levie2021radiounet, zhang2020cellular}. These methods predict signals by encoding and decoding images or point clouds of the scene. However, the encoded features fail to effectively capture the complex characteristics of the scene, resulting in large prediction errors. Recent advancements in multi-view 3D reconstruction introduce powerful tools for learning the radiance fields of scenes \cite{mildenhall2021nerf, kerbl20233d, huang20242d}. In the communication research, these techniques have been adapted for signal field modeling, yielding improved prediction accuracy \cite{zhao2023nerf2, wen2024wrf, lan2024acoustic}. Nevertheless, these methods typically assume that either the transmitter or receiver is fixed and learn signal distribution for the other’s movement. This assumption limits their ability to flexibly predict signal strength between arbitrary positions in a scene, as the signal field undergoes drastic changes when the fixed point shifts. In contrast, our method decomposes the signal propagation process into two explicit processes of direct paths and one implicit multipath process represented by a neural network. This approach enables flexible radio mapping with sparse data, allowing accurate prediction of signal strength for any given transmitter (Tx) and receiver (Rx) positions.

\subsection{Autonomous Radio Mapping}

Deploying mobile robots for autonomous environmental data collection and mapping is a critical application area \cite{Best-RSS-22, cao2023representation, dang2020graph, kratky2021autonomous}. For constructing radio maps, some methods rely on robots equipped with Wi-Fi access points to collect signal strength data, which is then used to model signal distribution. The simplest strategy, random sampling \cite{Clark-RSS-22}, is straightforward to implement but suffers from inefficiency due to redundant data collection in some regions and insufficient coverage in others. To address these limitations, uncertainty-driven sampling methods have been proposed \cite{banfi2017multirobot, fink2010online, amigoni2019online, penumarthi2017multirobot}. These approaches guide robots to areas with higher variance by fitting online models, such as Gaussian Processes, to the collected data. However, they rely on a single robot acting as a central planner to coordinate the team, which limits the efficiency of collaborative collection. Additionally, the computational resources required for model fitting further reduce the overall task efficiency. The authors of \cite{hsieh2004constructing} address the problem of radio map construction by formulating it as an optimal graph coverage problem. However, their approach requires a given communication graph and mainly focuses on solutions for two robots and offers a greedy strategy for three robots. In this paper, we first analyze the data requirements for multipath propagation modeling and introduce a visibility-based scene partitioning method to help meet these requirements. We then formulate the problem and propose a scalable solution for efficient autonomous data collection using large multi-robot teams. Our method enables efficient and flexible radio map construction in complex environments without any online fitted models, overcoming the limitations of existing approaches.

\section{Framework Overview}
The goal of this work is to develop an accurate and efficient framework for radio mapping in complex environments. We assume access to a 3D geometric map of the scene (e.g., a point cloud) and autonomous mobile robots capable of localizing within the map and following planned trajectories. The framework is designed to efficiently collect signal data and predict the received signal strength for any Tx-Rx pair. It consists of two core components: a hybrid radio propagation model for precise signal strength prediction (Fig. \ref{fig:pip}(a)) and a scalable autonomous data collection strategy utilizing a multi-robot system (Fig. \ref{fig:pip}(b)).

In Sec. \ref{sec:mapping}, we present our hybrid propagation modeling approach, which decomposes signal propagation into direct and multipath components, enabling effective generalization with sparse data. Based on the characteristics of this model, we define a data collection problem and propose a scalable planning strategy for multiple robots to collaboratively gather signal data (Sec. \ref{sec:collection}). The collected data is then used to train the hybrid model, enabling it to accurately predict signal strength at novel Tx-Rx positions.

\section{Hybrid Radio Propagation Modeling}
\label{sec:mapping}
FERMI is designed to predict the signal strength for arbitrary transmitter-receiver (Tx-Rx) pairs within a given scene. The key to this capability lies in dividing the signal propagation process into different stages: direct propagation and multipath propagation. To model the multipath propagation accurately, which is influenced by both the scene's geometric structure and the complex material properties, a data-driven approach is used, with neural networks learning the multipath propagation across the LOS points of Tx and Rx. Fig. \ref{fig:pip}(a) shows how we compute the propagation stages. We first calculate the signal from Tx to its LOS points through distance-related attenuation \cite{bahl2000radar}, which are then used as weights when summing their corresponding multipath propagation signals generated at each Rx LOS point. The positions of each LOS point of Tx and Rx are paired to form network inputs for calculating the multipath propagation signal and the attenuation coefficient for each Rx LOS point when its signal is transmitted to the Rx. Finally, we use a model similar to optical volume rendering \cite{max1995optical} to render the signals from all Rx LOS points into the signal received by the Rx and train the model using the collected measurements. 

Our hybrid architecture enables more effective learning of multipath propagation, rather than simply overfitting to the signal distribution of different Tx-Rx combinations. By training on sparsely sampled data points, our method can generalize to accurately predict signal strength for novel Tx-Rx pairs not included in the training set. The following sections detail how these modules are designed and trained.

\subsection{Multipath Propagation Network}

Given a geometric map of the scene (e.g., point clouds or occupancy grids), learning the signal strength contributed by multipath propagation presents two key problems:
\begin{itemize}
    \item \textbf{Generalization to unseen Tx-Rx pairs:} How can the multipath propagation network determine signal strength for novel Tx-Rx pairs beyond the training data?
    \item \textbf{Efficient training with sparse data:} How can the network effectively learn from limited, sparsely sampled data points?
\end{itemize}

To address the first challenge, we map each Tx-Rx pair to a set of LOS points in the scene that are relevant to the multipath propagation process. These LOS points are then used as inputs to the network. This transformation ensures that the network is agnostic to the specific positions of the Tx and Rx. As a result, the network can generalize to predict signal strength for unseen Tx-Rx pairs, provided their corresponding LOS points are covered by those in the training data. To tackle the second challenge, we design an efficient training architecture that allows a single data point to contribute to the training of multiple LOS points. This approach significantly reduces the need on densely sampled training data while maintaining prediction accuracy. The details of our method are presented in the following.

\subsubsection{Network inputs}

To enable the prediction of signal strength for Tx-Rx pairs beyond the training set, we design a network with position-agnostic inputs. This design allows the network to focus on capturing the multipath propagation characteristics of the environment, rather than being tied to specific Tx and Rx positions. Specifically, multipath propagation arises from the complex interactions of radio waves with scene surfaces. According to the Huygens-Fresnel principle, each point reached by a wave can be treated as a secondary source that retransmits the signal. Building on this principle, we decompose the multipath propagation process into three sequential stages:

\begin{enumerate}

    \item Direct path propagation from the transmitter to nearby visible surfaces.
    \item Signal propagation from the visible surfaces near Tx, acting as retransmitters, to the visible surfaces near Rx.
    \item Direct path propagation from visible surfaces near the Rx to the Rx.
    
\end{enumerate}

This decomposition is motivated by the observation that visible surfaces near the transmitter (Tx), which receive strong signals through direct paths, act as critical secondary sources. Similarly, visible surfaces near the receiver (Rx) significantly contribute to the received signal via direct radiation. To model the signal propagation, our approach explicitly computes the first and third stages using physical models, specifically the distance-related path attenuation model as introduced in \cite{bahl2000radar} and in \cite{max1995optical}. Meanwhile, the second stage, which captures the complex interactions between the signal and the environment, is modeled using a neural network. To effectively train the neural network for the second stage, it is crucial to construct a representation that efficiently encodes the structure of the scene. To achieve this, we map the Tx and Rx positions to a set of sampled LOS points, which are then used as inputs to the network. These points are generated by uniformly sampling rays on a sphere centered at the Tx and Rx positions and tracing their first hit points on the scene surfaces. The sampling process ensures that the neural network's input captures essential geometric information, while the network implicitly learns the material properties required to accurately model multipath propagation.

By using the LOS points of the Tx and Rx as inputs, the generalization capability of the multipath propagation module is greatly enhanced. This is because, even for previously unseen Tx-Rx pairs, their LOS points are largely present in the training set. By explicitly modeling the direct path propagation to these LOS points and implicitly modeling multipath propagation across different LOS points, our method effectively predicts signal strength even for unseen Tx-Rx pairs.

\subsubsection{Training Architecture}

We address the challenge of learning from sparse data by improving the utilization efficiency of each data point. Specifically, each data point contributes thousands of different network inputs during training. To achieve this, we treat each Tx LOS point as an independent retransmitter, where the received signal at an Rx LOS point is calculated by integrating the complex signals from all Tx LOS points. During the forward process, we iterate through the LOS points of Tx and Rx, selecting one point from each set at a time to form a single Tx-Rx LOS point pair as network input. For each Rx LOS point, the complex signals from all Tx LOS points are aggregated. These aggregated signals are then rendered into the final received signal at the Rx, which is then used for training with the collected data. This method effectively transforms a single Tx-Rx data point into thousands of sampled LOS points' pairs, significantly enhancing the utility of sparse data. Moreover, to address the ray cast errors caused by map noise, we perform additional sampling within a 0.5m range around the intersection of the Rx sampling rays and scene surface. The learned attenuation parameters implicitly account for the ray cast errors. The resulting LOS points of Tx and Rx are denoted as $\{T_j\}$ and $\{R_{w_ks_i}\}$ respectively, where $w_k$ denotes the sampling direction and $s_i$ denotes the index of the additional sampled points.

We adopt a network architecture inspired by \cite{zhao2023nerf2}, comprising two MLPs: an attenuation network and a radiance network. The attenuation network predicts the material-related attenuation coefficient at each Rx LOS point. The Rx LOS point's position $P_{R_{w_ks_i}} = (x,y,z)$ is fed into this fully connected network, producing an attenuation coefficient $\delta (P_{R_{w_ks_i}})$ and a feature vector. This feature vector is then concatenated with the position of a Tx LOS point $P_{T_j}$ and passed into the radiance network. The radiance network predicts the amplitude $a(P_{T_j},P_{R_{w_ks_i}})$ and phase $\theta(P_{T_j},P_{R_{w_ks_i}})$ of the complex signal retransmitted by the Tx LOS point to the Rx LOS point. The positions of both Tx and Rx LOS points are encoded using hash features \cite{muller2022instant}. Formally, the multipath propagation network $\mathcal{F}$ is represented as follows:

\begin{equation}
    \mathcal{F}_{\Theta}: (P_{T_j},P_{R_{w_ks_i}}) \rightarrow \left( \delta (P_{R_{w_ks_i}}),S(P_{T_j},P_{R_{w_ks_i}}) \right)
\end{equation}

where $\Theta$ indicates the learnable parameters and $S(P_{T_j},P_{R_{w_ks_i}})=a(P_{T_j},P_{R_{w_ks_i}})e^{j\theta(P_{T_j},P_{R_{w_ks_i}})}$. 

\subsection{Radio Strength Rendering}

\begin{figure}[t]
    \centering
    \includegraphics[width=0.9\linewidth, trim=1mm 1mm 1mm 1mm, clip]{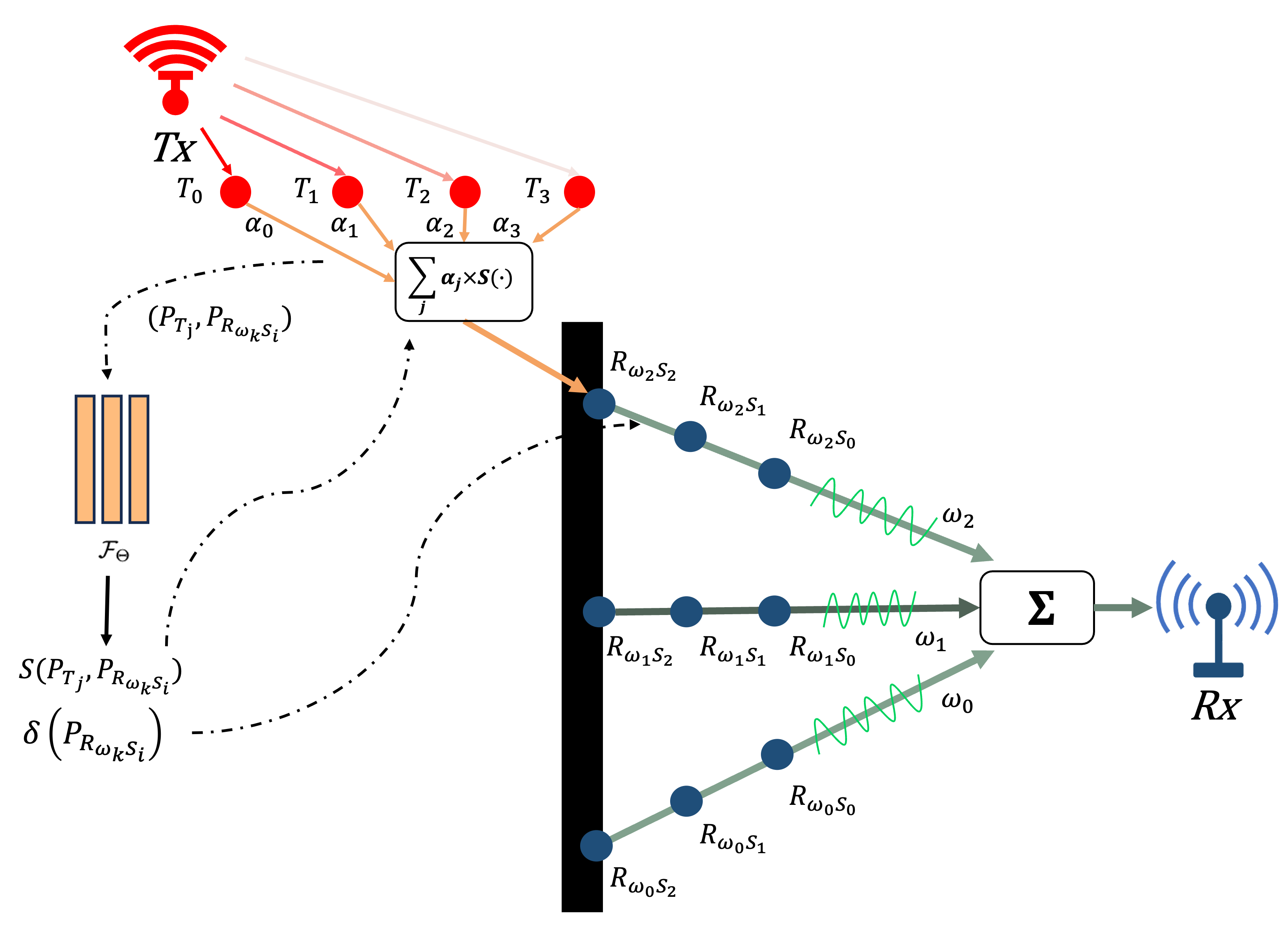}
    \caption{An illustration demonstrates the rendering process for multipath propagation. Each LOS point of the Tx acts as a retransmitter. The energy propagated from the Tx to a LOS point via direct path propagation determines the weight assigned to that point when combining signals at the corresponding LOS point of the Rx. After calculating the signals transmitted through multipath propagation between each pair of Tx and Rx LOS points via the network, the received signals at each Rx LOS point are combined with the weights of the corresponding Tx LOS points. The signal contribution of each ray is then computed by integrating along its path. Finally, the signals from all rays are summed to determine the total received signal through multipath propagation.} 
    \vspace{-3mm}
    \label{fig:render}
\end{figure}

After computing the complex signal from each Tx LOS point, we perform a weighted integration at each Rx LOS point:

\begin{equation}
S(P_{R_{w_ks_i}}) = \sum_{j} \alpha_j \cdot S(P_{T_j},P_{R_{w_ks_i}})),
\end{equation}

where $\alpha_j$ is the normalized energy of the direct path attenuation from the Tx to the $P_{T_j}$. Once the integrated signal for each $P_{R_{w_ks_i}}$ is obtained, we accumulate the signals along each Rx ray sample. This process is similar to volume rendering for light \cite{max1995optical}, where the network-predicted $\delta (P_{R_{w_ks_i}})$ are used to accumulate signal along the ray:

\begin{equation}
h(w) = \sum_{i} L_{R_{w_ks_i}}(1-exp(\sigma_{R_{w_ks_i}}\delta( P_{R_{w_ks_i}})))S(P_{R_{w_ks_i}})
\end{equation}

where $\sigma_{R_{w_ks_i}}$ equals the distance between $P_{R_{w_ks_i}}$ and $P_{R_{w_ks_{i+1}}}$, and:

\begin{equation}
L_{R_{w_ks_i}}=exp(-\sum_{n=1}^{i-1}\sigma_{R_{w_ks_n}}\delta (P_{R_{w_ks_n}}))
\end{equation}

The final received signal by the Rx is computed by integrating the signal across all ray directions $w$, weighted by the receiver's gain pattern $G(w)$:

\begin{equation}
R_{Rx} = \sum_{k} G(w_k) h(\omega_k)
\end{equation}

In our application, we use an omnidirectional antenna, where the gain function $G(\omega)$ is equal for all directions. Fig. \ref{fig:render} illustrates the computational process of rendering for multipath propagation.

The previous part computes the signal contribution of multipath propagation. However, under LOS conditions, the direct path propagation from Tx to Rx often dominates over multipath propagation. We calculate the weighted sum of the direct path energy and multipath energy in these cases as follows:
\begin{equation}
    \mathcal{R}_{Rx}= \alpha_{los} \cdot R_{ref} +(1-\alpha_{los}) \cdot |R_{Rx}|
\end{equation}

Here $R_{ref}$ denotes the reference signal strength of the Tx, while $\alpha_{los}$ denotes the distance-related attenuation factor associated with direct path loss under LOS conditions and is equal to 0 under NLOS conditions.

\subsection{Optimization}

We supervise the network using measured signal strength data $RSSI_{Rx}$:
\begin{equation}
    \mathcal{L}_{energy}=|\mathcal{R}_{Rx} - RSSI_{Rx}|^2
\end{equation}

Additionally, to mitigate the abnormally large signal values produced by Tx LOS points and prevent numerical instability after integration, we introduce an energy decay loss:
\begin{equation}
    \mathcal{L}_{decay} = \sum |S(P_{T_j},P_{R_{w_ks_i}})|
\end{equation}

Our total loss is a linear combination of the above loss:
\begin{equation}
    \mathcal{L}_{total} = \lambda_1\mathcal{L}_{energy} + \lambda_2\mathcal{L}_{decay}
\end{equation}

\section{Scalable autonomous data collection }  
\label{sec:collection}

Training the multipath propagation network requires considerable data collection, which traditionally involves deploying Wi-Fi access nodes across various positions in the environment. To address this challenge, we propose an autonomous data collection strategy using a team of mobile robots equipped with Wi-Fi access nodes. This approach eliminates the need for labor-intensive manual deployment. Furthermore, since each Wi-Fi node carried by the robots can function simultaneously as both a transmitter and a receiver, the autonomous system enables parallel data collection across multiple regions.

We first analyze the data requirements of the multipath propagation network. Then, we propose a data collection method that effectively meets these requirements. To autonomous collect the data efficiently, we define and solve the problem of cooperative data collection using multiple robots.

To comprehensively train the multipath propagation network, the dataset must capture sufficient information about multipath signals between mutually non-visible surface points. To achieve this, every surface point should be paired with all of its non-visible surface points as input for training. While this requires a large amount of input data, our framework addresses this challenge by leveraging each collected Tx-Rx pair to generate multiple combinations of surface point pairs. This enables sparse Tx-Rx signal data to effectively cover multiple mutually non-visible points. As a result, if the LOS points from all Tx-Rx pairs collectively cover all mutually non-visible points, the dataset will provide the necessary information for comprehensive network training.

% To construct such a dataset, we propose a visibility-based scene partitioning method that enables sufficient data collection. Specifically, we cluster scene surfaces based on mutual visibility, grouping mutually visible surfaces into the same region. We then deploy Tx-Rx pairs between every two regions to collect data. This approach ensures that the collected dataset satisfies the requirement of LOS points covering non-visible surfaces. Since the non-visible surfaces of any region are completely covered by other regions, deploying Wi-Fi nodes in each region while iterating through all others ensures that the collected LOS points effectively cover all non-visible surfaces. Consequently, this guarantees sufficient training data for the network. The scene partitioning method is detailed in \ref{space patition sec}.

To construct such a dataset, we propose a visibility-based scene partitioning method that facilitates efficient data collection. Specifically, we cluster scene surfaces based on mutual visibility, grouping mutually visible surface points into the same region. Tx-Rx pairs are then deployed between every two regions to collect data. This strategy ensures that each region’s surface points are paired with those from other regions, covering possible propagation paths. Since the visibility-based partitioning ensures comprehensive coverage of the scene, every possible interaction between surface points is captured through these Tx-Rx pairs. This method guarantees that all necessary multipath components are included in the dataset, ensuring the network is trained on a wide variety of propagation scenarios. Consequently, the dataset construction strategy is sufficient for training a robust and generalized multipath propagation model. The scene partitioning method is detailed in Sec. \ref{space patition sec}.

Building on the scene partitioning method, the data collection task is formalized as ensuring that Wi-Fi nodes are deployed across all possible pairs of regions. To efficiently manage this task, we translate it into a collection state matrix and provide a formal problem definition in Sec. \ref{definition}. Moreover, efficient solutions tailored to various robot team sizes are presented in Sec. \ref{solution}, ensuring scalability and practicality for real-world deployments.

\subsection{Problem Definition}
\label{definition}
The set of the regions after scene partitioning is represented as $V$. A total of $n$ robots equipped with Wi-Fi nodes are deployed across $n$ regions to collect pairwise data, including Tx's position, Rx's position and signal strength. To monitor the progress of data collection, we define a collection state matrix $M$, where rows and columns correspond to the regions. Initially, the diagonal elements of $M$ are set to 1, indicating that data collection within the same region is not required, while all other elements are initialized to 0, indicating that data collection between those regions is incomplete. When Wi-Fi nodes deployment between regions $i$ and $j$ is completed, the corresponding matrix element $M(i,j)$ is updated to 1, indicating that data collection between these two region is complete. If communication coverage between $i$ and $j$ is not achievable, the element is updated to -1. Since robots in the ad-hoc network function as both transmitters and receivers, the collection state matrix $M$ is symmetric, satisfying $M(i,j) = M(j,i)$.

\begin{figure}[t]
    \centering
    \includegraphics[width=0.9\linewidth, trim=2mm 2mm 3mm 3mm, clip]{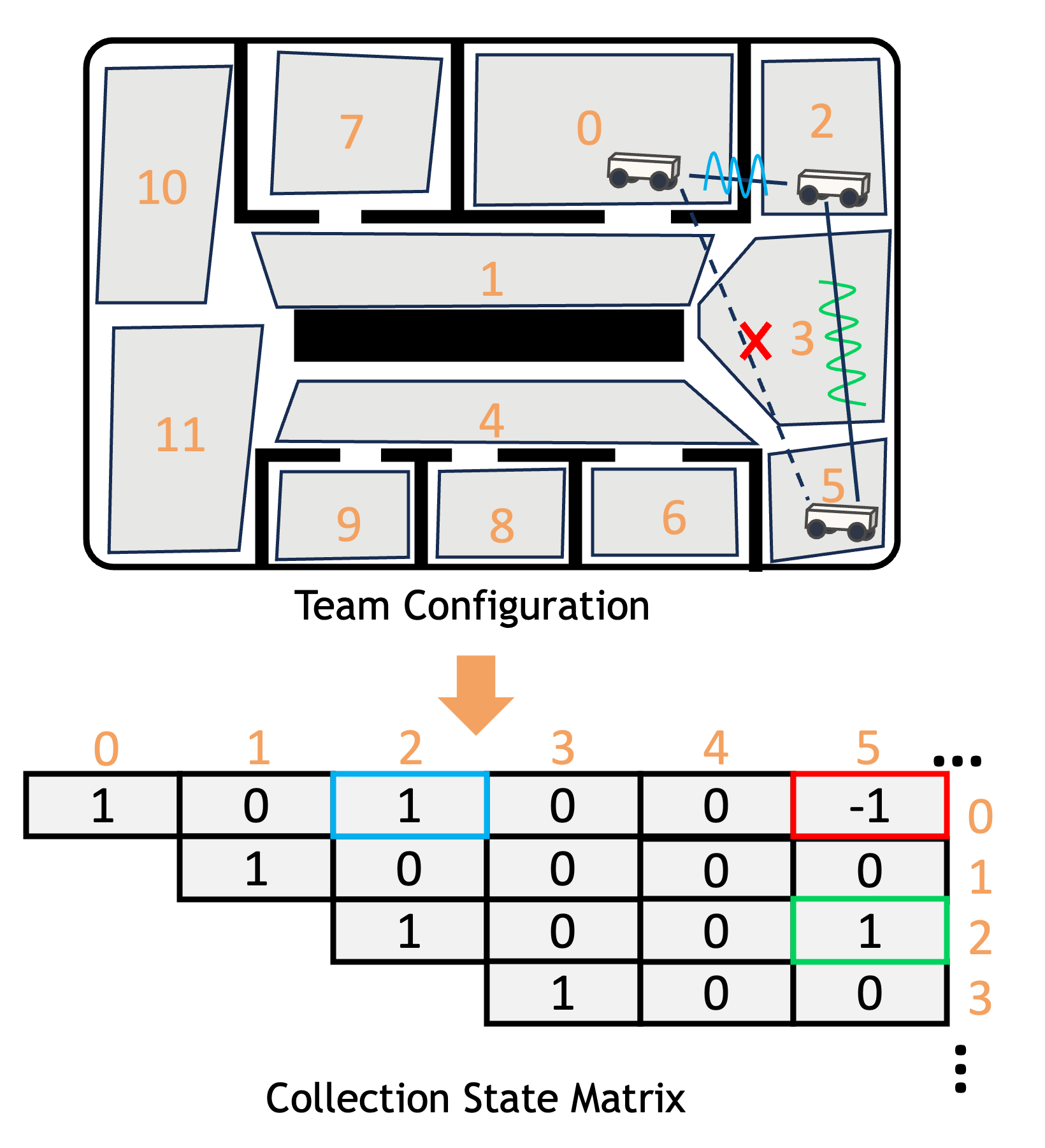}
    \caption{An illustration showing the robot team configuration and how the corresponding collection state matrix elements are updated.} 
    \vspace{-3mm}
    \label{fig:transision}
\end{figure}

For parallel data collection with $n$ robots, the set of regions occupied by the robots is denoted as $\mathcal{C}=\{v_p \mid v_p \in V, p=0, \dots, n-1\}$, referred to as the team configuration. Region pairs $S_{\mathcal{C}}=\{(v_p, v_q) \mid p, q = 0, \dots, n-1, p \neq q\}$ are mapped to elements in the collection state matrix $M$. Once all robots complete the deployment, data collection for this configuration is considered complete, and the corresponding matrix elements are updated to 1. A schematic illustrating the relationship between collection state matrix updates and robot configurations is shown in Fig. \ref{fig:transision}. After completing data collection for the current configuration, the robot team transitions to the next configuration, continues data collection, and updates the corresponding elements in $M$. The process repeats until all elements of $M$ are either $1$, indicating completion, or $-1$, indicating infeasibility. At this point, the data collection process is considered complete. Our objective is to find a sequence of configurations to maximize parallelism in data collection and minimize transition costs between the configuration sequence, as detailed in the next subsection.

\subsection{Scalable Solution}
\label{solution}
To efficiently complete the data collection process defined above, the deployment strategy focuses on two key aspects:
\begin{itemize}
    \item \textbf{Parallelism:} To maximize the parallelism capacity of the $n$-robot configuration, each transitioned configuration should ideally correspond to elements in $M$ that are 0, avoiding deployment in previously collected region pairs.
    \item \textbf{Minimal transition cost:} To minimize time wasted during transitions between configurations, the total transition cost of the team should be minimized. The transition cost between any two configurations is defined as the cost of the optimal bipartite matching of their robot positions.
\end{itemize}

We discuss two cases to satisfy the above objectives:

\mypara{Case 1: $n=2$}
In this case, each configuration occupies two regions, corresponding directly to two symmetric elements in $M$. The set of all configurations corresponds to all elements in $M$ except those on the diagonal. Starting from an initial configuration, the configuration sequence minimizing transition costs can be solved as a Traveling Salesman Problem (TSP) on the configuration set \cite{helsgaun2000effective}.

\mypara{Case 2: $n>2$}
Here, each $n$-robot configuration covers $\frac{n(n-1)}{2}$ elements of $M$, making the optimal configuration sequence a combinatorial optimization problem which is NP-hard. To address this, we use a two-step hierarchical approach:

\begin{enumerate}
    \item \textbf{Parallelism Maximization:} We formulate the problem as a set coverage problem. Given a universal set $U$:
    \begin{equation}
        U=\{(u,v)\mid u,v\in V,u\neq v\}
    \end{equation} 
    representing uncollected region pairs, and $N$ subsets $S_{\mathcal{C}}$, we find the minimal subset collection $\{S_{\mathcal{C}}\}$ that covers $U$. This ensures maximum parallelism and minimizes the number of transitions. Since the size of $\{S_{\mathcal{C}}\}$ grows factorially with $n$, we adopt a greedy algorithm for practical implementation.
    
    \item \textbf{Transition Cost Minimization:} After identifying the minimal configuration set, we solve a TSP on the set, starting from the initial configuration, to obtain the sequence with the lowest transition cost.
\end{enumerate}

This strategy allows robot teams of various sizes to maximize parallel data collection while minimizing transition costs. For the case where $n > 2$, the solution to the set coverage problem may not be unique, potentially leading to suboptimal results. However, practical experiments demonstrate that the obtained solutions are acceptable for real-world applications. In real-world data collection scenarios, if two regions do not exist communication coverage, the robot team reverts to the previous configuration and re-solves the problem using the remaining uncollected region pairs as $U$. This ensures that the data collection process can continue despite communication limitations.

\begin{figure}[t]
    \centering
    \includegraphics[width=0.9\linewidth, trim=2mm 2mm 3mm 3mm, clip]{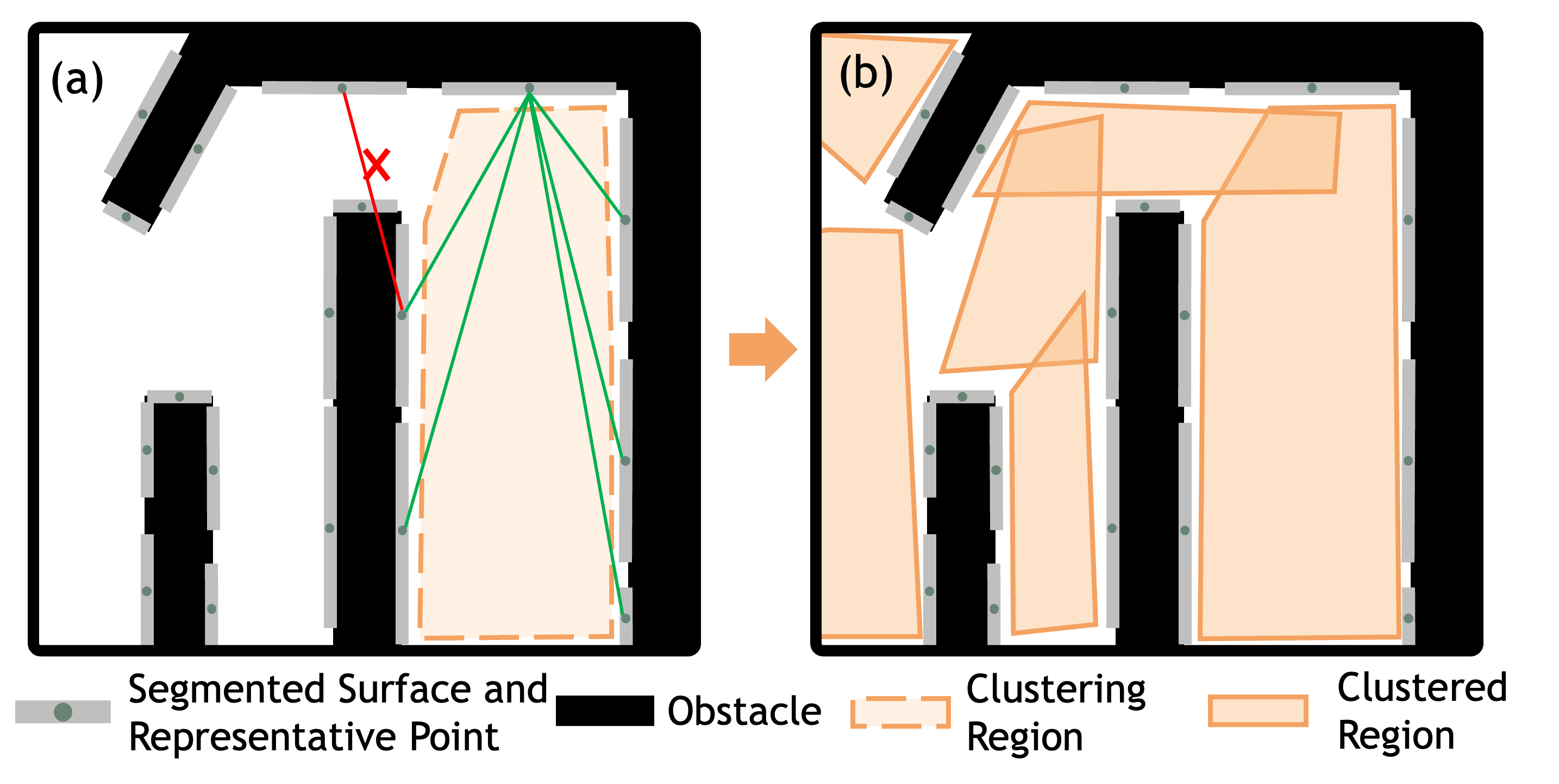}
    \caption{A schematic for our visibility-based scene partitioning method. (a) After segmenting the scene's surface and obtaining representative points, we cluster each representative surface point. A point is added to a cluster if it has visibility to all surface points already in that cluster. If it lacks visibility to any point in the cluster, it will not be added. (b) Illustration of the final partitioning results.} 
    \vspace{-3mm}
    \label{fig:cluster}
\end{figure}

\subsection{Visibility-based Scene Partitioning}
\label{space patition sec}
We divide the scene so that all surface points within each region are mutually visible. To reduce computation costs prior to partitioning, the 3D scene is first voxelized to discretize the space into a uniform grid of voxels. Surface voxels are then grouped based on euclidean distance to form connected components. To prevent any single group from becoming excessively large, PCA-based segmentation is applied: for large surface regions, the covariance matrix of the 3D positions of the voxels is computed, followed by eigenvalue decomposition, and the surface is split along the direction of the largest eigenvalue (principal component). After segmentation, the average center point of each segmented surface is computed and selected as its representative point. These representative points are used in the subsequent scene partitioning process, reducing computational complexity while retaining essential structural information.

Using these representative points, we perform clustering. The process begins with a randomly chosen representative point. For each cluster, an unclustered point is added if it is mutually visible with all points in the current cluster and its distance to the cluster center is less than a threshold $D$. Otherwise, the algorithm moves to the next unclustered point closest to the current cluster center, repeating the process until no valid unclustered points remain. Once a cluster is complete, the next clustering iteration begins from the unclustered point closest to the last cluster center. An illustration of the clustering process is provided in Fig. \ref{fig:cluster}.

\begin{figure}[t]
    \centering
    \includegraphics[width=0.98\linewidth]{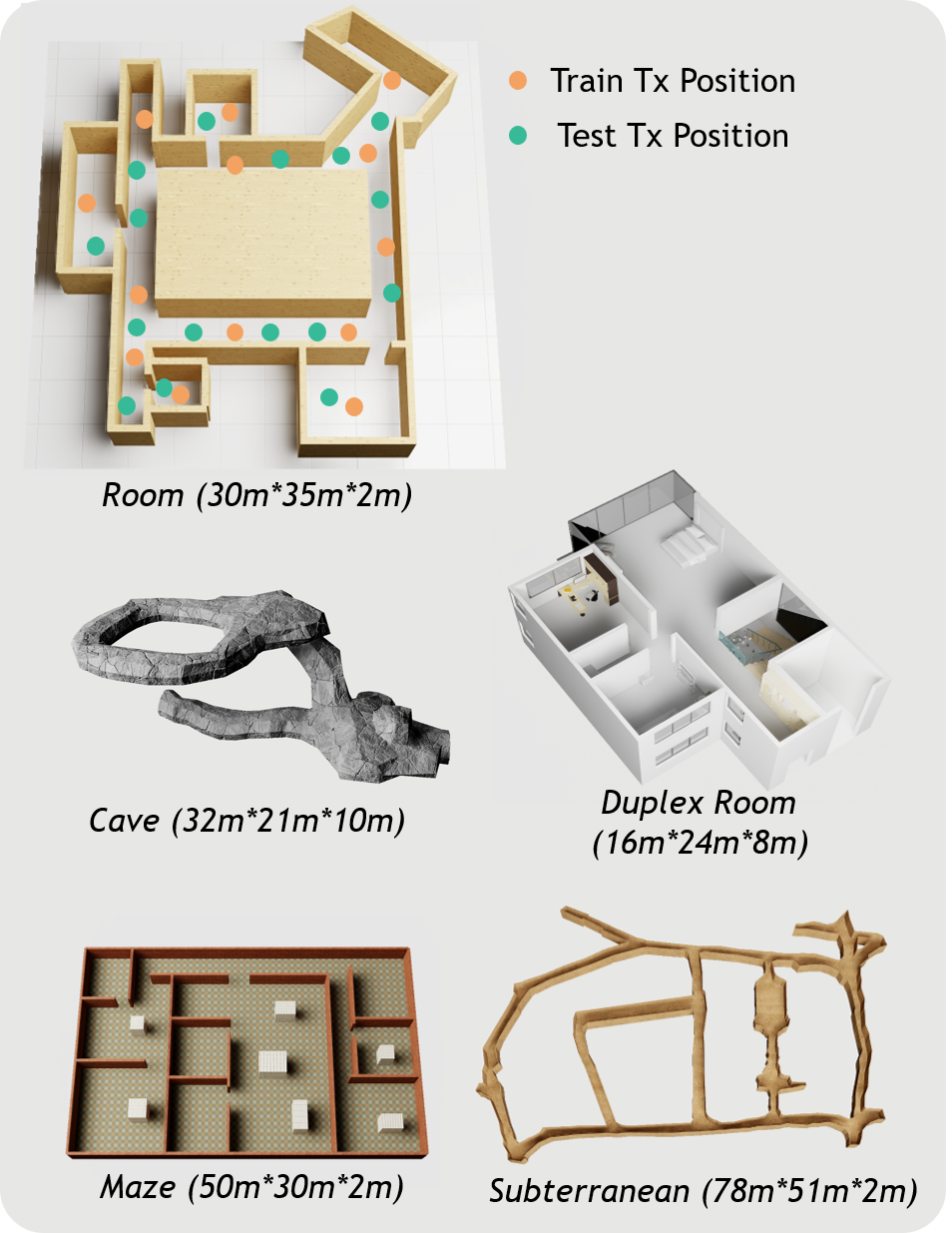}
    \caption{The simulation experiment scenarios and a schematic of the training and testing Tx in the room scene.} 
    \vspace{-3mm}
    \label{fig:exp_scene}
\end{figure}

\begin{figure*}[t] 
    \centering
    \includegraphics[width=\linewidth, clip, trim=2mm 2mm 2mm 2mm]{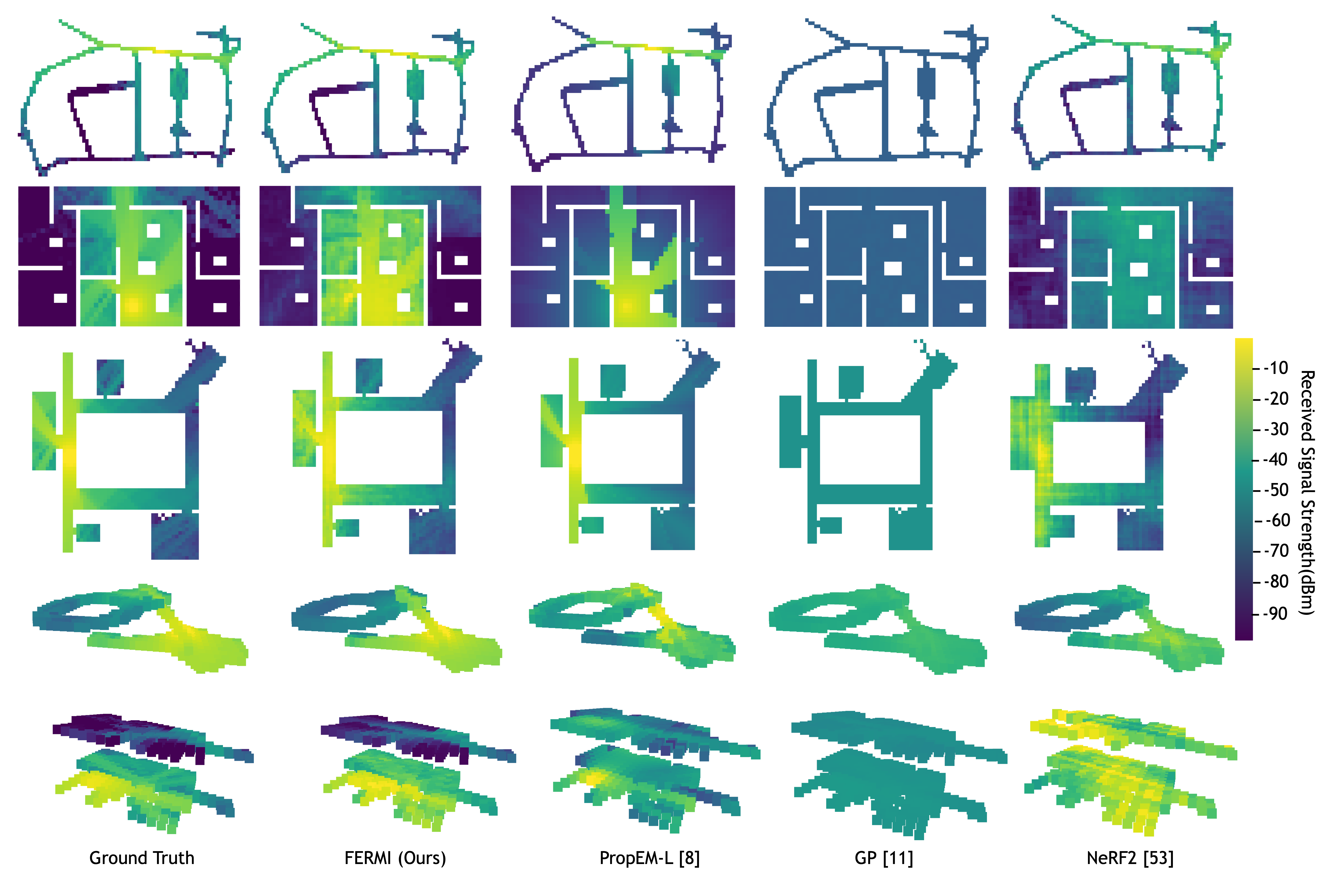}
    \caption{Visualization of spatial signal data generated by different methods in each simulation scenario. Ground truth data is produced by a fixed testing Tx.}
    \label{fig:exp1}
\end{figure*}

\section{Experiments}

In our experiment, we first evaluate the performance of the proposed FERMI framework from two aspects, comparing it with state-of-the-art (SOTA) methods:

\begin{itemize}
    \item \textbf{Generalization:} We assess the model's ability to predict signal strength for unseen Tx-Rx pairs, verifying its generalization capability across arbitrary Tx-Rx positions.
    \item \textbf{Data utilization efficiency:} We evaluate the model's performance when trained on increasingly sparse datasets, demonstrating its ability to achieve accurate radio mapping with limited training data.
\end{itemize}

To validate generalization, the model is trained on sparsely sampled data and evaluated on comprehensively sampled Tx-Rx pairs from the entire environment. To assess data utilization efficiency, we progressively reduce the size of the training data and measure the prediction error on the same test dataset.
Additionally, to evaluate the efficiency of autonomous data collection, we compare our method with existing approaches under varying numbers of robots, using metrics including the number of transitions, overall transition costs and computation time. Finally, we deploy a team of three robots in a real-world environment to autonomously collect data and train the model, further validating its practical applicability.

\subsection{Simulation Results}
\subsubsection{Quantitative Results for Signal Strength Prediction}
We compare our methods with existing radio mapping methods for signal strength prediction on simulation datasets across multiple scenarios. Specifically, we use Blender software to create five 3D maps of environments with complex occlusions and generate signal data using the Sionna ray tracing engine \cite{sionna} with default texture and material parameters. To ensure sufficient training data coverage, each scene is divided into several regions using our partitioning method (Sec. \ref{space patition sec}), and a random point within each region is selected as the transmitter (Tx) position. For each Tx, we generate signal strength data at a resolution of 1 meters. The data from all Txs are then combined to construct the training dataset.

To evaluate the generalization capability of each method for predicting signal strength at unseen positions, we randomly sample multiple Tx positions across the entire scene that are not included in the training set and generate signal strength data at the same resolution to construct the test dataset. Fig. \ref{fig:exp_scene} illustrates the 3D models of all five scenes and shows the sampled training and testing Tx positions in one specific scene.

We compare the performance of the following three existing methods:

\begin{enumerate}
    \item \textbf{PropEM-L \cite{Clark-RSS-22}:} This method encodes scene information using a geometric map of the environment and decodes the signal strength of Tx-Rx pairs at different locations through a neural network.    
    \item \textbf{Gaussian Processes (GP) \cite{fink2010online}:} This method uses Gaussian Processes to fit and predict the signal strength of Tx-Rx pairs at different locations.
    \item \textbf{NeRF2 \cite{zhao2023nerf2}:} A neural radiance field (NeRF) based method to model the radio radiance field.
\end{enumerate}

\begin{table}[t]
\caption{\Revise{MAE of Predicted Signal Strength (dBm)}}
\label{TAB:Exp1}
\resizebox{\linewidth}{!}
{
\begin{tabular}{cccccc}

\hline
         & Room          & Cave          & Maze           & Subterranean  & \begin{tabular}[c]{@{}l@{}}Duplex\\  Room\end{tabular} \\
\hline
FERMI (Ours)     & \textbf{5.59} & \textbf{5.49} & \textbf{10.52} & \textbf{6.98} & \textbf{11.29}                                         \\
PropEM-L \cite{Clark-RSS-22} & 9.12          & 9.73          & 16.32          & 16.49         & 18.79                                                  \\
GP \cite{fink2010online}       & 18.18         & 17.52         & 25.03          & 19.14         & 25.88                                                  \\
NeRF2 \cite{zhao2023nerf2}    & 10.09         & 9.27          & 17.10          & 11.61         & 14.30   \\
\hline
\end{tabular}
}
\end{table}

In the comparisons, we evaluate the performance of each method using the mean absolute error (MAE) between the predicted signal strength and the ground truth. The results are summarized in Table. \ref{TAB:Exp1}, which shows that our proposed method consistently achieves the lowest MAE across all scenarios, with improvements of up to 40\%. Furthermore, we visualize the predicted signal distributions for a specific test Tx position. The visualization results in Fig. \ref{fig:exp1} demonstrate that our method accurately captures the spatial discontinuities in signal strength distribution, especially in cases of rapid signal attenuation caused by occlusions and multipath effects resulting from multiple interactions between the signal and the environment.

\begin{figure}[t]
    \centering
    \includegraphics[width=0.98\linewidth]{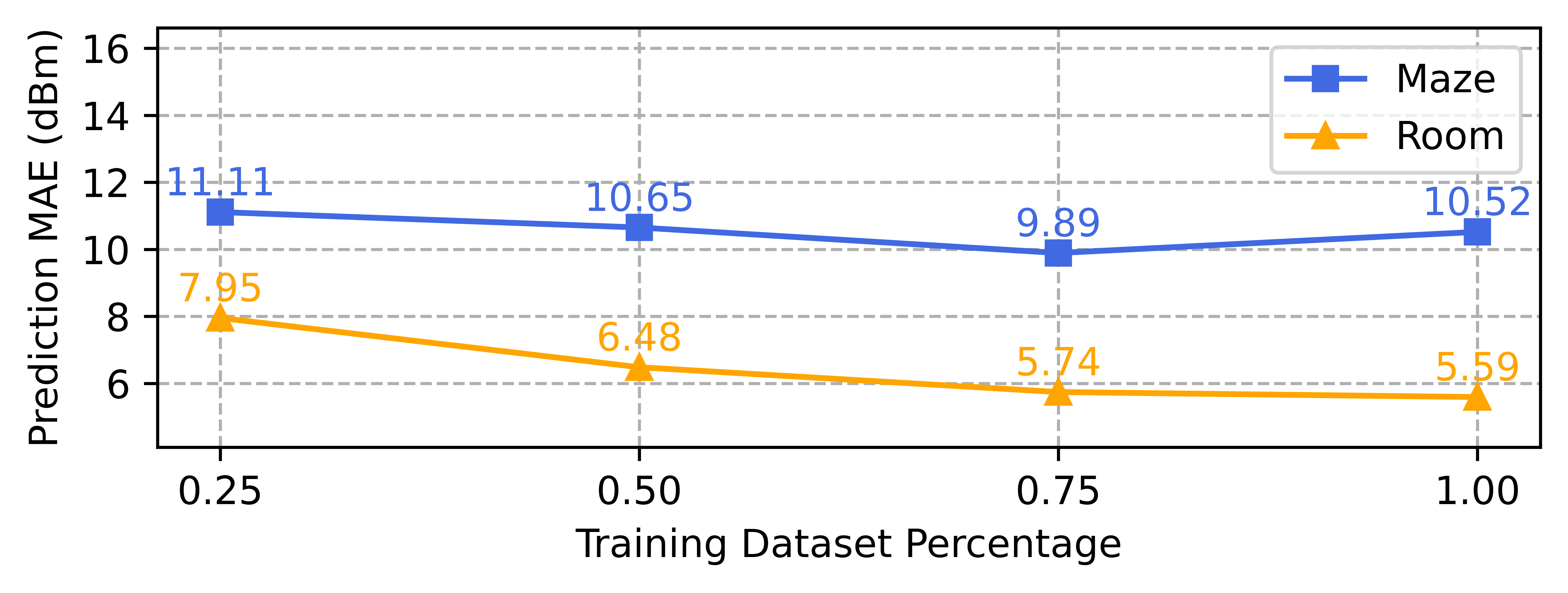}
    \caption{The mean absolute errors for signal prediction corresponding to different percentages of reduction in the training set.} 
    \label{fig:abla_1}
    \vspace{-5mm}
\end{figure}

\subsubsection{Ablation Studies}
To validate the data utilization efficiency of our method and assess the effectiveness of its module designs, we conduct the following ablation experiments:

\underline{\textbf{Training with Datasets of Different Sizes}} Our method achieves significantly lower errors than benchmark approaches on the original training set. To further assess its data utilization efficiency, we evaluate its performance when the training data is reduced. We evaluate two scenarios and progressively reduce the percentage of training data to sparsify the dataset. The model is retrained on each sparsified dataset and evaluated using the corresponding test data. The results are shown in Fig. \ref{fig:abla_1}.

\begin{figure}[t]
    \centering
    \includegraphics[width=0.98\linewidth,clip, trim=3mm 3mm 6mm 6mm]{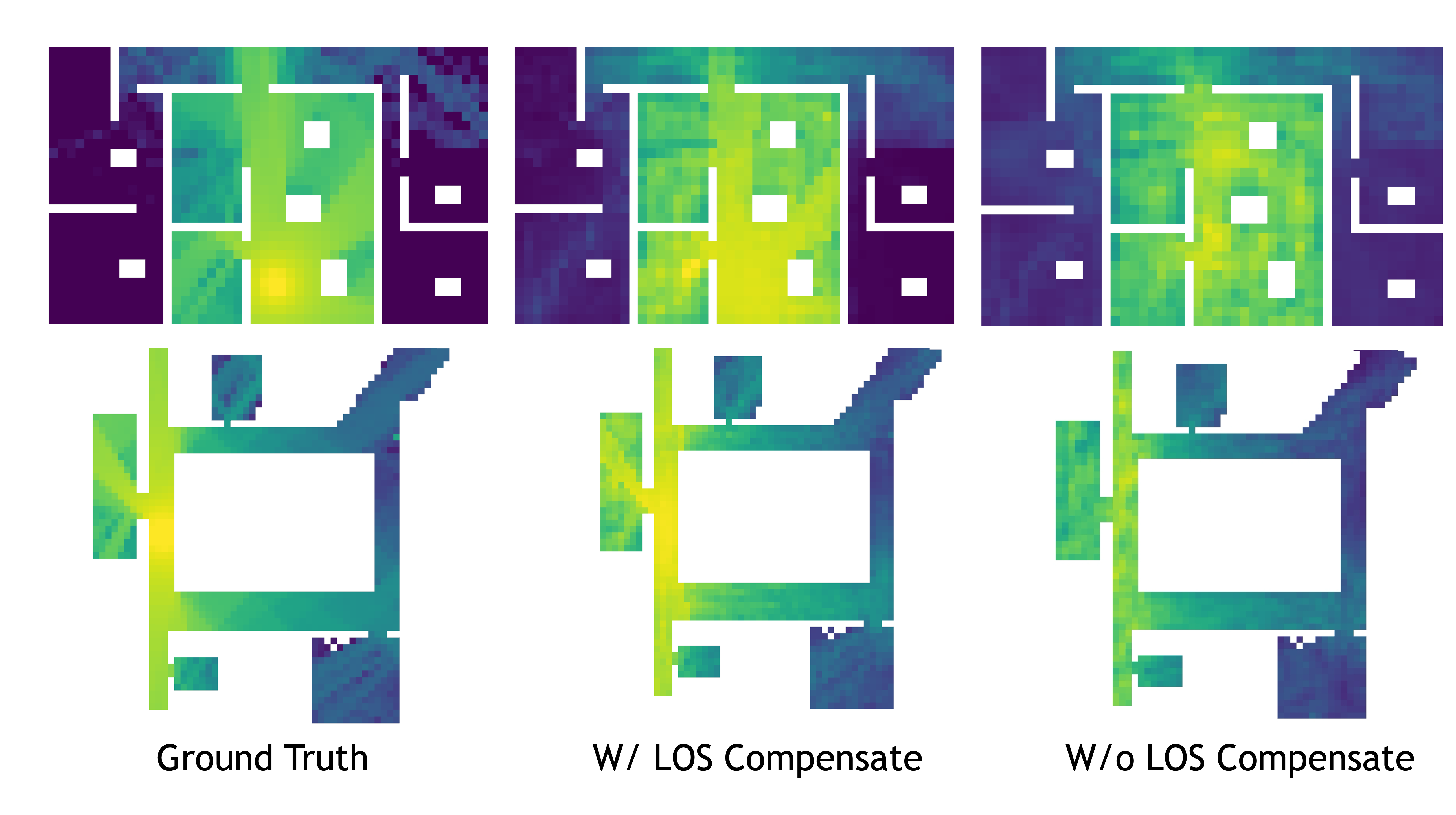}
    \caption{A comparison of prediction results with and without Line-of-Sight (LOS) signal strength compensation in our method.} 
    \vspace{-3mm}
    \label{fig:abla_2}
\end{figure}

The experimental results demonstrate that even with extremely sparse training data, our method shows only minor degradation in prediction accuracy, and continues to outperform baseline methods that use more data. This highlights the data utilization efficiency of our approach, indicating that it can achieve satisfactory prediction results even when trained on highly sparse datasets. 

\underline{\textbf{Strength Compensation under LOS Condition}} In line-of-sight (LOS) scenarios, the direct path attenuation of the signal dominates the total received signal strength. Therefore, we propose a compensation technique to adjust the predicted strength. The effectiveness of this technique is validated in two scenarios, with results shown in Fig. \ref{fig:abla_2}. The results indicate that without this compensation, our method tends to underestimate the total signal strength in LOS conditions, as the multipath signals account for only part of the total strength. By applying the compensation, the prediction accuracy improves significantly, demonstrating the necessity of this technique.

\subsubsection{Quantitative Results for Autonomous Data Collection Planning}
To evaluate the efficiency of our planning strategy for autonomous signal data collection, we compare it with a previous method \cite{hsieh2004constructing}. This method employs a greedy strategy for three robots, selecting the configuration that increases new measurements while minimizing transition cost at each step. Additionally, it requires a given communication graph, where the edges and vertices indicate which areas in the scene need Wi-Fi node deployment for signal collection. For a comprehensive comparison, we convert our collection matrix into the communication graph required by this method and extend its greedy strategy to accommodate multiple robots. Experiments are conducted in the maze scene, which is divided into 22 regions. As a result, the data collection matrix contains 231 elements that need to be updated.

Table \ref{TAB:Collection} presents the number of configuration transitions, the corresponding transition costs, and the total computation time for both our method and the benchmark method \cite{hsieh2004constructing} under varying numbers of robots. The results indicate that our planning strategy is highly scalable, maintaining planning times within a few seconds as the number of robots increases. In contrast, the benchmark method requires searching through numerous configurations at each step to identify the transition that minimizes cost while increasing measurements. As the number of robots grows, this leads to a significant computation burden. Additionally, our method effectively reduces unnecessary configuration transitions and transition costs by leveraging parallel cooperation between robots, thereby optimizing task performance.

\begin{table}[t]
\caption{\Revise{Quantitative Results of Autonomous Data Collection}}
\label{TAB:Collection}
\resizebox{\linewidth}{!}
{
\begin{tabular}{@{}cccc@{}}
\toprule
                 & Transition Number & Transition Cost (meter) & Computation Time (second) \\ \midrule
Ours-3 Robots    & 88                 & 1464.73   & 0.07          \\
Ours-9 Robots    & 10                & 356.34    & 5.59           \\
Ours-15 Robots   & 3                 & 148.57    & 1.44          \\
\midrule
Greedy\cite{hsieh2004constructing}-3 Robots  & 208               & 2228.53  & 1.12              \\
Greedy\cite{hsieh2004constructing}-9 Robots  & 105               & 1125.26  & 721.24            \\
Greedy\cite{hsieh2004constructing}-15 Robots & 36                & 449.28   & 205.38           \\
\bottomrule
\end{tabular}
}
\vspace{-7mm}
\end{table}

\subsection{Real-World Experiments}
\subsubsection{Platform Setup}

We use three Agile X Scout Mini robots as our experimental platform. Each robot is equipped with a Livox Mid-360 Lidar for perception and uses the method proposed in \cite{xu2021fast} to construct point cloud maps and occupancy grid maps. For onboard computation, each robot is equipped with an NUC i7 computer, while ad-hoc network cards are used as Wi-Fi access nodes.

Given a set of target region points, each robot employs the trajectory optimization method and MPC controller described in \cite{wu2024real}. Communication between robots is implemented using a TCP-based system \cite{zhou2021ego}, enabling the sharing of each robot’s odometry and signal strength data.

\subsubsection{Results}

\begin{figure}[t]
    \centering
    \includegraphics[width=0.98\linewidth,clip, trim=3mm 3mm 6mm 6mm]{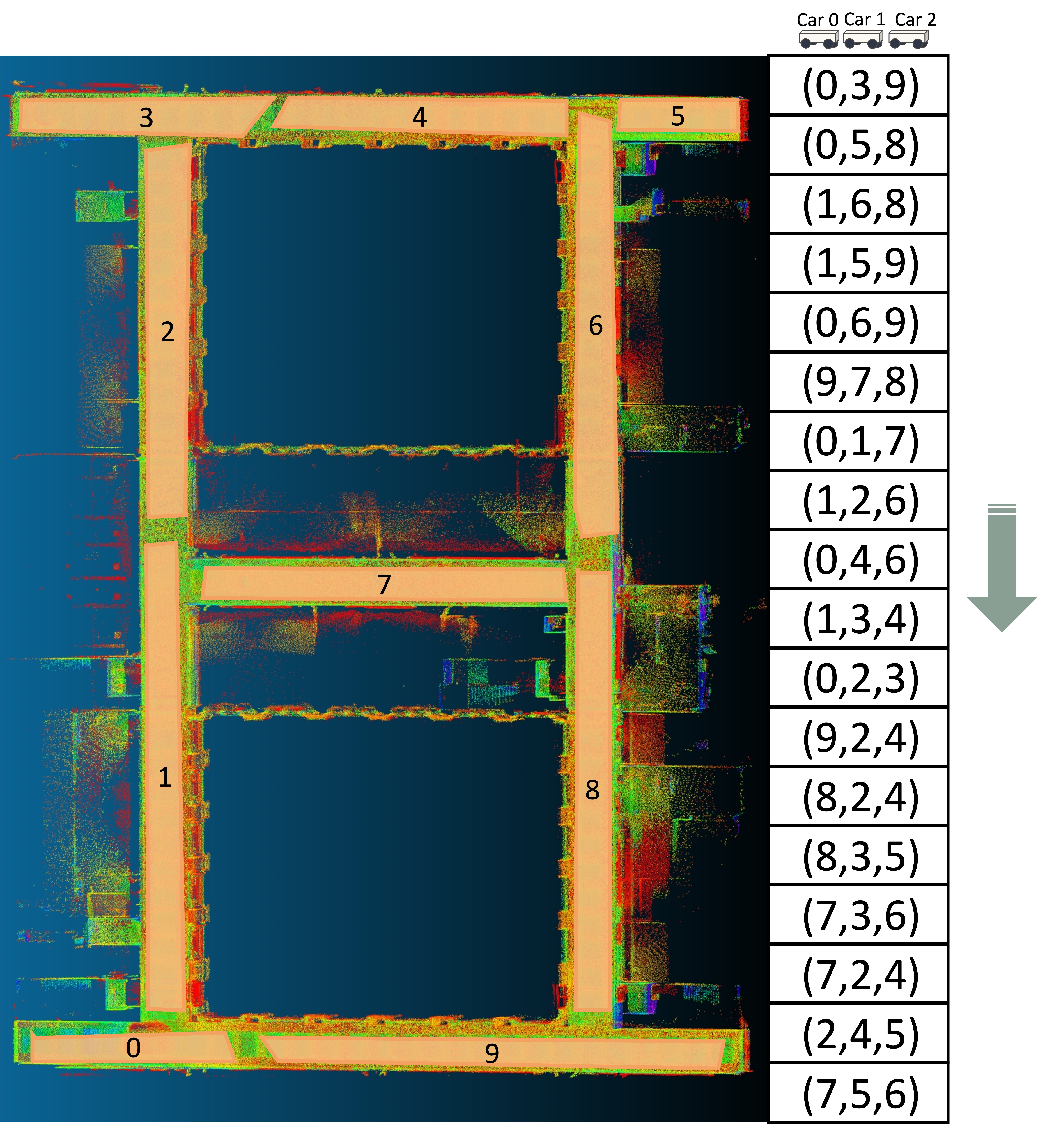}
    \caption{The scene partitioning result and the configuration sequence of the three robots.} 
    \label{fig:real1}
    \vspace{-3mm}
\end{figure}

\begin{figure}[t]
    \centering
    \includegraphics[width=0.98\linewidth,clip, trim=1mm 1mm 2mm 2mm]{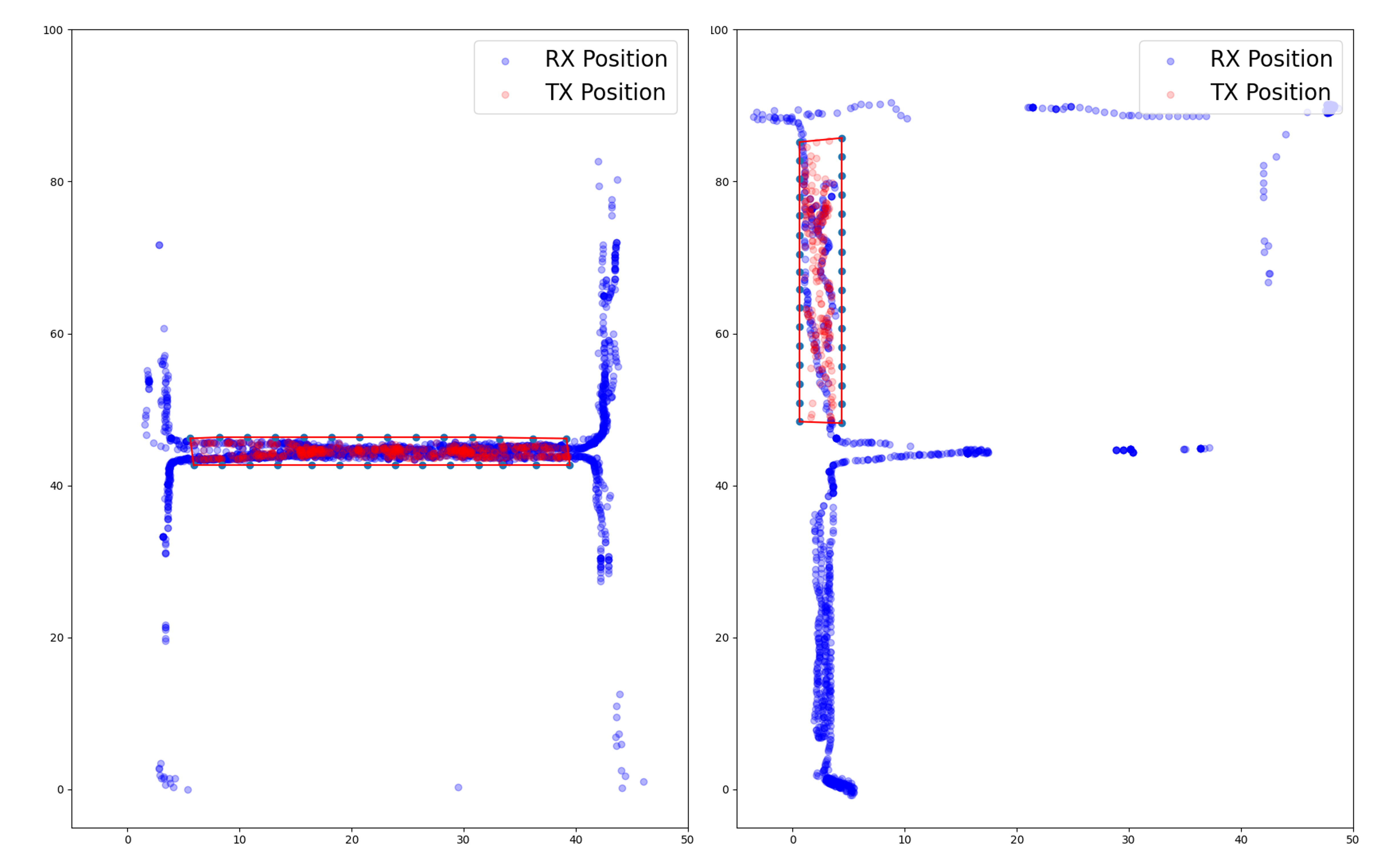}
    \caption{Examples for collected signal data.} 
    \label{fig:real2}
    \vspace{-3mm}
\end{figure}

We deploy three robots in an indoor corridor environment bounded by 94m × 64m × 3m to perform signal data collection. The point clouds and a photograph of the environment are provided in Fig. \ref{fig:teaser}. The scene partitioning result and the configuration sequence is shown as Fig. \ref{fig:real1}. Our method collect comprehensive signal data without human intervention within 1 hour, resulting in a dataset containing around 40,000 valid data. Fig. \ref{fig:real2} shows signal data collected when the Txs are deployed in two region respectively. To address noisy fluctuations of the signal during collection, we apply median filtering to nearby data points. For regions with no signal measurements, we perform data augmentation using the methods described in \cite{Clark-RSS-22, miyagusuku2016improving}, filling the signal strength with the minimum measured value (-80 dBm in our experiment). 

The collected data is randomly split into 50\% for training and 50\% for testing, with the training data ensuring coverage between every pair of regions. The training data is used to train our method alongside PropEM-L \cite{Clark-RSS-22} and NeRF2 \cite{zhao2023nerf2}, and the mean absolute error (MAE) of predictions is computed on the test set. The results, presented in Table. \ref{TAB: real}, indicate that our method achieves the lowest MAE.

\begin{table}
    \centering
    \caption{\Revise{Results of Real-world Signal Prediction}}
    \begin{tabular}{c|c}
         & Test Set MAE (dBm) \\
        \hline
        Ours & 6.24 \\
        PropEM-L \cite{Clark-RSS-22} & 8.52 \\
        NeRF2 \cite{zhao2023nerf2} & 9.15 \\
    \end{tabular}
    \vspace{-7mm}
    \label{TAB: real}
\end{table}

Additionally, we collect dense signal data generated from a fixed Tx point to perform qualitative analysis. As shown in Fig. \ref{fig:teaser}, the visualization demonstrates that our method effectively captures the discontinuous distribution of the signal, whereas the predictions from other methods appear overly smooth.

\section{Limitations and Future Works}
\label{Limitations}

Although FERMI provides a flexible and accurate framework for radio mapping, a few limitations remain. First, our method assumes a static environment, limiting its applicability in dynamic scenes. Future work could explore the impact of scene layout changes on signal propagation.

Second, the approach requires training for each individual scene. Future efforts could focus on encoding scene information to enable generalization across unseen environments.

Third, our method requires a point cloud map of the scene for scene partitioning and subsequent data collection planning. Future work could consider planning methods that integrate data collection with exploration task, enabling simultaneous execution.

\section{Conclusion} 
\label{sec:conclusion}

In this work, we propose FERMI, a flexible radio mapping framework capable of accurately predicting signal strength in large-scale environments with complex occlusions. Our proposed hybrid propagation model divides signal propagation into two explicit physical processes and one implicit process represented by a neural network that captures multipath propagation. This design enables signal prediction between any two points in the scene using sparse signal data. Additionally, our scalable planning method can be deployed on robot teams of varying sizes, enabling autonomous signal data collection. This reduces the need for extensive human labor while improving parallel efficiency  through robot collaboration. We hope that our radio mapping framework provides an effective tool for communication-aware robot team collaboration.

\section*{Acknowledgments}
This work was supported by the Young Scientists Fund of the National Natural Science Foundation of China (Grant No. 62403502
). We would like to thank Zitong Lan, Qianyue He, Xiaopeng Zhao for their thoughtful discussions.

% \section*{Acknowledgments}

%% Use plainnat to work nicely with natbib. 

\bibliographystyle{plainnat}
\bibliography{references}

\begin{thebibliography}{55}
\providecommand{\natexlab}[1]{#1}
\providecommand{\url}[1]{\texttt{#1}}
\expandafter\ifx\csname urlstyle\endcsname\relax
  \providecommand{\doi}[1]{doi: #1}\else
  \providecommand{\doi}{doi: \begingroup \urlstyle{rm}\Url}\fi

\bibitem[Amigoni et~al.(2019)Amigoni, Banfi, Basilico, Rekleitis, and
  Quattrini~Li]{amigoni2019online}
Francesco Amigoni, Jacopo Banfi, Nicola Basilico, Ioannis Rekleitis, and
  Alberto Quattrini~Li.
\newblock Online update of communication maps for exploring multirobot systems
  under connectivity constraints.
\newblock In \emph{Distributed Autonomous Robotic Systems: The 14th
  International Symposium}, pages 513--526. Springer, 2019.

\bibitem[Bahl and Padmanabhan(2000)]{bahl2000radar}
Paramvir Bahl and Venkata~N Padmanabhan.
\newblock Radar: An in-building rf-based user location and tracking system.
\newblock In \emph{Proceedings IEEE INFOCOM 2000. Conference on computer
  communications. Nineteenth annual joint conference of the IEEE computer and
  communications societies (Cat. No. 00CH37064)}, volume~2, pages 775--784.
  Ieee, 2000.

\bibitem[Banfi et~al.(2017)Banfi, Li, Basilico, Rekleitis, and
  Amigoni]{banfi2017multirobot}
Jacopo Banfi, Alberto~Quattrini Li, Nicola Basilico, Ioannis Rekleitis, and
  Francesco Amigoni.
\newblock Multirobot online construction of communication maps.
\newblock In \emph{2017 IEEE International Conference on Robotics and
  Automation (ICRA)}, pages 2577--2583. IEEE, 2017.

\bibitem[Banfi et~al.(2018)Banfi, Basilico, and Carpin]{banfi2018optimal}
Jacopo Banfi, Nicola Basilico, and Stefano Carpin.
\newblock Optimal redeployment of multirobot teams for communication
  maintenance.
\newblock In \emph{2018 IEEE/RSJ International Conference on Intelligent Robots
  and Systems (IROS)}, pages 3757--3764. IEEE, 2018.

\bibitem[Best et~al.(2022)Best, Garg, Keller, Hollinger, and
  Scherer]{Best-RSS-22}
Graeme Best, Rohit Garg, John Keller, {Geoffrey A.} Hollinger, and Sebastian
  Scherer.
\newblock {Resilient Multi-Sensor Exploration of Multifarious Environments with
  a Team of Aerial Robots}.
\newblock In \emph{Proceedings of Robotics: Science and Systems}, New York
  City, NY, USA, June 2022.
\newblock \doi{10.15607/RSS.2022.XVIII.004}.

\bibitem[Campos et~al.(2023)Campos, Cid, Vangasse, J{\'u}nior, Bai{\~a}o,
  Pimenta, Barros, Pessin, and Freitas]{campos2023navigation}
Sofia~Pereira Campos, Andr{\'e} Luiz~Maciel Cid, Arthur Da~Costa Vangasse,
  Gilmar J{\'u}nior, Jo{\~a}o Bai{\~a}o, Luciano Pimenta, Luiz Barros, Gustavo
  Pessin, and Gustavo Freitas.
\newblock Navigation strategy with multi-robots in confined environments to
  improve radio signal coverage.
\newblock In \emph{2023 Latin American Robotics Symposium (LARS), 2023
  Brazilian Symposium on Robotics (SBR), and 2023 Workshop on Robotics in
  Education (WRE)}, pages 313--318. IEEE, 2023.

\bibitem[Cao et~al.(2023)Cao, Zhu, Ren, Choset, and
  Zhang]{cao2023representation}
Chao Cao, Hongbiao Zhu, Zhongqiang Ren, Howie Choset, and Ji~Zhang.
\newblock Representation granularity enables time-efficient autonomous
  exploration in large, complex worlds.
\newblock \emph{Science Robotics}, 8\penalty0 (80):\penalty0 eadf0970, 2023.

\bibitem[Clark et~al.(2022)Clark, Edlund, Net, Vaquero, and akbar
  Agha-mohammadi]{Clark-RSS-22}
Lillian Clark, Jeffrey Edlund, {Marc Sanchez} Net, {Tiago Stegun} Vaquero, and
  Ali akbar Agha-mohammadi.
\newblock {PropEM-L: Radio Propagation Environment Modeling and Learning for
  Communication-Aware Multi-Robot Exploration}.
\newblock In \emph{Proceedings of Robotics: Science and Systems}, New York
  City, NY, USA, June 2022.
\newblock \doi{10.15607/RSS.2022.XVIII.014}.

\bibitem[Dang et~al.(2020)Dang, Tranzatto, Khattak, Mascarich, Alexis, and
  Hutter]{dang2020graph}
Tung Dang, Marco Tranzatto, Shehryar Khattak, Frank Mascarich, Kostas Alexis,
  and Marco Hutter.
\newblock Graph-based subterranean exploration path planning using aerial and
  legged robots.
\newblock \emph{Journal of Field Robotics}, 37\penalty0 (8):\penalty0
  1363--1388, 2020.

\bibitem[Dixon and Frew(2009)]{dixon2009maintaining}
Cory Dixon and Eric~W Frew.
\newblock Maintaining optimal communication chains in robotic sensor networks
  using mobility control.
\newblock \emph{Mobile Networks and Applications}, 14:\penalty0 281--291, 2009.

\bibitem[Fink and Kumar(2010)]{fink2010online}
Jonathan Fink and Vijay Kumar.
\newblock Online methods for radio signal mapping with mobile robots.
\newblock In \emph{2010 IEEE International Conference on Robotics and
  Automation}, pages 1940--1945. IEEE, 2010.

\bibitem[Fink et~al.(2013)Fink, Ribeiro, and Kumar]{fink2013robust}
Jonathan Fink, Alejandro Ribeiro, and Vijay Kumar.
\newblock Robust control of mobility and communications in autonomous robot
  teams.
\newblock \emph{IEEE Access}, 1:\penalty0 290--309, 2013.

\bibitem[Guo and Zavlanos(2018)]{guo2018multirobot}
Meng Guo and Michael~M Zavlanos.
\newblock Multirobot data gathering under buffer constraints and intermittent
  communication.
\newblock \emph{IEEE transactions on robotics}, 34\penalty0 (4):\penalty0
  1082--1097, 2018.

\bibitem[H{\"a}hnel and Fox(2006)]{hahnel2006gaussian}
Brian Ferris~Dirk H{\"a}hnel and Dieter Fox.
\newblock Gaussian processes for signal strength-based location estimation.
\newblock In \emph{Proceeding of robotics: science and systems}. Citeseer,
  2006.

\bibitem[Helsgaun(2000)]{helsgaun2000effective}
Keld Helsgaun.
\newblock An effective implementation of the lin--kernighan traveling salesman
  heuristic.
\newblock \emph{European journal of operational research}, 126\penalty0
  (1):\penalty0 106--130, 2000.

\bibitem[Hoydis et~al.(2022)Hoydis, Cammerer, {Ait Aoudia}, Vem, Binder,
  Marcus, and Keller]{sionna}
Jakob Hoydis, Sebastian Cammerer, Fayçal {Ait Aoudia}, Avinash Vem, Nikolaus
  Binder, Guillermo Marcus, and Alexander Keller.
\newblock Sionna: An open-source library for next-generation physical layer
  research.
\newblock \emph{arXiv preprint}, Mar. 2022.

\bibitem[Hoydis et~al.(2023)Hoydis, Aoudia, Cammerer, Nimier-David, Binder,
  Marcus, and Keller]{hoydis2023sionna}
Jakob Hoydis, Fay{\c{c}}al~A{\"\i}t Aoudia, Sebastian Cammerer, Merlin
  Nimier-David, Nikolaus Binder, Guillermo Marcus, and Alexander Keller.
\newblock Sionna rt: Differentiable ray tracing for radio propagation modeling.
\newblock In \emph{2023 IEEE Globecom Workshops (GC Wkshps)}, pages 317--321.
  IEEE, 2023.

\bibitem[Hoydis et~al.(2024)Hoydis, Aoudia, Cammerer, Euchner, Nimier-David,
  Ten~Brink, and Keller]{hoydis2024learning}
Jakob Hoydis, Fay{\c{c}}al~A{\"\i}t Aoudia, Sebastian Cammerer, Florian
  Euchner, Merlin Nimier-David, Stephan Ten~Brink, and Alexander Keller.
\newblock Learning radio environments by differentiable ray tracing.
\newblock \emph{IEEE Transactions on Machine Learning in Communications and
  Networking}, 2024.

\bibitem[Hsieh et~al.(2004)Hsieh, Kumar, and Taylor]{hsieh2004constructing}
M-YA Hsieh, Vijay Kumar, and Camillo~J Taylor.
\newblock Constructing radio signal strength maps with multiple robots.
\newblock In \emph{IEEE International Conference on Robotics and Automation,
  2004. Proceedings. ICRA'04. 2004}, volume~4, pages 4184--4189. IEEE, 2004.

\bibitem[Huang et~al.(2024)Huang, Yu, Chen, Geiger, and Gao]{huang20242d}
Binbin Huang, Zehao Yu, Anpei Chen, Andreas Geiger, and Shenghua Gao.
\newblock 2d gaussian splatting for geometrically accurate radiance fields.
\newblock In \emph{ACM SIGGRAPH 2024 conference papers}, pages 1--11, 2024.

\bibitem[Kerbl et~al.(2023)Kerbl, Kopanas, Leimk{\"u}hler, and
  Drettakis]{kerbl20233d}
Bernhard Kerbl, Georgios Kopanas, Thomas Leimk{\"u}hler, and George Drettakis.
\newblock 3d gaussian splatting for real-time radiance field rendering.
\newblock \emph{ACM Trans. Graph.}, 42\penalty0 (4):\penalty0 139--1, 2023.

\bibitem[Kotaru et~al.(2015)Kotaru, Joshi, Bharadia, and
  Katti]{kotaru2015spotfi}
Manikanta Kotaru, Kiran Joshi, Dinesh Bharadia, and Sachin Katti.
\newblock Spotfi: Decimeter level localization using wifi.
\newblock In \emph{Proceedings of the 2015 ACM conference on special interest
  group on data communication}, pages 269--282, 2015.

\bibitem[Kr{\'a}tk{\`y} et~al.(2021)Kr{\'a}tk{\`y}, Petr{\'a}{\v{c}}ek,
  B{\'a}{\v{c}}a, and Saska]{kratky2021autonomous}
V{\'\i}t Kr{\'a}tk{\`y}, Pavel Petr{\'a}{\v{c}}ek, Tom{\'a}{\v{s}}
  B{\'a}{\v{c}}a, and Martin Saska.
\newblock An autonomous unmanned aerial vehicle system for fast exploration of
  large complex indoor environments.
\newblock \emph{Journal of field robotics}, 38\penalty0 (8):\penalty0
  1036--1058, 2021.

\bibitem[Kulkarni et~al.(2022)Kulkarni, Dharmadhikari, Tranzatto, Zimmermann,
  Reijgwart, De~Petris, Nguyen, Khedekar, Papachristos, Ott,
  et~al.]{kulkarni2022autonomous}
Mihir Kulkarni, Mihir Dharmadhikari, Marco Tranzatto, Samuel Zimmermann, Victor
  Reijgwart, Paolo De~Petris, Huan Nguyen, Nikhil Khedekar, Christos
  Papachristos, Lionel Ott, et~al.
\newblock Autonomous teamed exploration of subterranean environments using
  legged and aerial robots.
\newblock In \emph{2022 International Conference on Robotics and Automation
  (ICRA)}, pages 3306--3313. IEEE, 2022.

\bibitem[Lan et~al.(2024)Lan, Zheng, Zheng, and Zhao]{lan2024acoustic}
Zitong Lan, Chenhao Zheng, Zhiwei Zheng, and Mingmin Zhao.
\newblock Acoustic volume rendering for neural impulse response fields.
\newblock \emph{arXiv preprint arXiv:2411.06307}, 2024.

\bibitem[Lee and Molisch(2024)]{lee2024scalable}
Ju-Hyung Lee and Andreas~F Molisch.
\newblock A scalable and generalizable pathloss map prediction.
\newblock \emph{IEEE Transactions on Wireless Communications}, 2024.

\bibitem[Levie et~al.(2021)Levie, Yapar, Kutyniok, and
  Caire]{levie2021radiounet}
Ron Levie, {\c{C}}a{\u{g}}kan Yapar, Gitta Kutyniok, and Giuseppe Caire.
\newblock Radiounet: Fast radio map estimation with convolutional neural
  networks.
\newblock \emph{IEEE Transactions on Wireless Communications}, 20\penalty0
  (6):\penalty0 4001--4015, 2021.

\bibitem[Lu et~al.(2024)Lu, Vattheuer, Mirzasoleiman, and Abari]{lu2024deep}
Haofan Lu, Christopher Vattheuer, Baharan Mirzasoleiman, and Omid Abari.
\newblock A deep learning framework for wireless radiation field reconstruction
  and channel prediction.
\newblock \emph{arXiv preprint arXiv:2403.03241}, 2024.

\bibitem[Max(1995)]{max1995optical}
Nelson Max.
\newblock Optical models for direct volume rendering.
\newblock \emph{IEEE Transactions on Visualization and Computer Graphics},
  1\penalty0 (2):\penalty0 99--108, 1995.

\bibitem[Mildenhall et~al.(2021)Mildenhall, Srinivasan, Tancik, Barron,
  Ramamoorthi, and Ng]{mildenhall2021nerf}
Ben Mildenhall, Pratul~P Srinivasan, Matthew Tancik, Jonathan~T Barron, Ravi
  Ramamoorthi, and Ren Ng.
\newblock Nerf: Representing scenes as neural radiance fields for view
  synthesis.
\newblock \emph{Communications of the ACM}, 65\penalty0 (1):\penalty0 99--106,
  2021.

\bibitem[Miyagusuku et~al.(2016)Miyagusuku, Yamashita, and
  Asama]{miyagusuku2016improving}
Renato Miyagusuku, Atsushi Yamashita, and Hajime Asama.
\newblock Improving gaussian processes based mapping of wireless signals using
  path loss models.
\newblock In \emph{2016 IEEE/RSJ International Conference on Intelligent Robots
  and Systems (IROS)}, pages 4610--4615. IEEE, 2016.

\bibitem[M{\"u}ller et~al.(2022)M{\"u}ller, Evans, Schied, and
  Keller]{muller2022instant}
Thomas M{\"u}ller, Alex Evans, Christoph Schied, and Alexander Keller.
\newblock Instant neural graphics primitives with a multiresolution hash
  encoding.
\newblock \emph{ACM transactions on graphics (TOG)}, 41\penalty0 (4):\penalty0
  1--15, 2022.

\bibitem[Orekondy et~al.(2023)Orekondy, Kumar, Kadambi, Ye, Soriaga, and
  Behboodi]{orekondy2023winert}
Tribhuvanesh Orekondy, Pratik Kumar, Shreya Kadambi, Hao Ye, Joseph Soriaga,
  and Arash Behboodi.
\newblock Winert: Towards neural ray tracing for wireless channel modelling and
  differentiable simulations.
\newblock In \emph{The Eleventh International Conference on Learning
  Representations}, 2023.

\bibitem[Penumarthi et~al.(2017)Penumarthi, Li, Banfi, Basilico, Amigoni,
  O'Kane, Rekleitis, and Nelakuditi]{penumarthi2017multirobot}
Phani~Krishna Penumarthi, Alberto~Quattrini Li, Jacopo Banfi, Nicola Basilico,
  Francesco Amigoni, Jason O'Kane, Ioannis Rekleitis, and Srihari Nelakuditi.
\newblock Multirobot exploration for building communication maps with prior
  from communication models.
\newblock In \emph{2017 International Symposium on Multi-Robot and Multi-Agent
  Systems (MRS)}, pages 90--96. IEEE, 2017.

\bibitem[Ruah et~al.(2024)Ruah, Simeone, Hoydis, and
  Al-Hashimi]{ruah2024calibrating}
Clement Ruah, Osvaldo Simeone, Jakob Hoydis, and Bashir Al-Hashimi.
\newblock Calibrating wireless ray tracing for digital twinning using local
  phase error estimates.
\newblock \emph{IEEE Transactions on Machine Learning in Communications and
  Networking}, 2024.

\bibitem[Sabattini et~al.(2013)Sabattini, Chopra, and
  Secchi]{sabattini2013decentralized}
Lorenzo Sabattini, Nikhil Chopra, and Cristian Secchi.
\newblock Decentralized connectivity maintenance for cooperative control of
  mobile robotic systems.
\newblock \emph{The International Journal of Robotics Research}, 32\penalty0
  (12):\penalty0 1411--1423, 2013.

\bibitem[Schwaighofer et~al.(2003)Schwaighofer, Grigoras, Tresp, and
  Hoffmann]{schwaighofer2003gpps}
Anton Schwaighofer, Marian Grigoras, Volker Tresp, and Clemens Hoffmann.
\newblock Gpps: A gaussian process positioning system for cellular networks.
\newblock \emph{Advances in Neural Information Processing Systems}, 16, 2003.

\bibitem[Shin et~al.(2014)Shin, Chon, Kim, and Cha]{shin2014mri}
Hyojeong Shin, Yohan Chon, Yungeun Kim, and Hojung Cha.
\newblock Mri: Model-based radio interpolation for indoor war-walking.
\newblock \emph{IEEE Transactions on Mobile Computing}, 14\penalty0
  (6):\penalty0 1231--1244, 2014.

\bibitem[Stump et~al.(2008)Stump, Jadbabaie, and Kumar]{stump2008connectivity}
Ethan Stump, Ali Jadbabaie, and Vijay Kumar.
\newblock Connectivity management in mobile robot teams.
\newblock In \emph{2008 IEEE international conference on robotics and
  automation}, pages 1525--1530. IEEE, 2008.

\bibitem[Stump et~al.(2011)Stump, Michael, Kumar, and
  Isler]{stump2011visibility}
Ethan Stump, Nathan Michael, Vijay Kumar, and Volkan Isler.
\newblock Visibility-based deployment of robot formations for communication
  maintenance.
\newblock In \emph{2011 IEEE international conference on robotics and
  automation}, pages 4498--4505. IEEE, 2011.

\bibitem[Tardioli et~al.(2010)Tardioli, Mosteo, Riazuelo, Villarroel, and
  Montano]{tardioli2010enforcing}
Danilo Tardioli, Alejandro~R Mosteo, Luis Riazuelo, Jos{\'e}~Luis Villarroel,
  and Luis Montano.
\newblock Enforcing network connectivity in robot team missions.
\newblock \emph{The International Journal of Robotics Research}, 29\penalty0
  (4):\penalty0 460--480, 2010.

\bibitem[Tekdas et~al.(2010)Tekdas, Yang, and Isler]{tekdas2010robotic}
Onur Tekdas, Wei Yang, and Volkan Isler.
\newblock Robotic routers: Algorithms and implementation.
\newblock \emph{The International Journal of Robotics Research}, 29\penalty0
  (1):\penalty0 110--126, 2010.

\bibitem[Tian et~al.(2024)Tian, Zhang, Wei, and Guo]{tian2024ihero}
Zhuoli Tian, Yuyang Zhang, Jinsheng Wei, and Meng Guo.
\newblock ihero: Interactive human-oriented exploration and supervision under
  scarce communication.
\newblock \emph{arXiv preprint arXiv:2405.12571}, 2024.

\bibitem[Wen et~al.(2024)Wen, Tong, Hu, Lin, and Zhang]{wen2024wrf}
Chaozheng Wen, Jingwen Tong, Yingdong Hu, Zehong Lin, and Jun Zhang.
\newblock Wrf-gs: Wireless radiation field reconstruction with 3d gaussian
  splatting.
\newblock \emph{arXiv preprint arXiv:2412.04832}, 2024.

\bibitem[Wu et~al.(2024)Wu, Wang, Song, Gao, Mei, and Zhou]{wu2024real}
Chengkai Wu, Ruilin Wang, Mianzhi Song, Fei Gao, Jie Mei, and Boyu Zhou.
\newblock Real-time whole-body motion planning for mobile manipulators using
  environment-adaptive search and spatial-temporal optimization.
\newblock In \emph{2024 IEEE International Conference on Robotics and
  Automation (ICRA)}, pages 1369--1375. IEEE, 2024.

\bibitem[Xia et~al.(2023)Xia, Deng, Pan, Zhang, Duan, Zhou, and
  Cheng]{xia2023relink}
Lijia Xia, Beiming Deng, Jie Pan, Xiaoxun Zhang, Peiming Duan, Boyu Zhou, and
  Hui Cheng.
\newblock Relink: Real-time line-of-sight-based deployment framework of
  multi-robot for maintaining a communication network.
\newblock \emph{IEEE Robotics and Automation Letters}, 2023.

\bibitem[Xu and Zhang(2021)]{xu2021fast}
Wei Xu and Fu~Zhang.
\newblock Fast-lio: A fast, robust lidar-inertial odometry package by
  tightly-coupled iterated kalman filter.
\newblock \emph{IEEE Robotics and Automation Letters}, 6\penalty0 (2):\penalty0
  3317--3324, 2021.

\bibitem[Yan and Mostofi(2012)]{yan2012robotic}
Yuan Yan and Yasamin Mostofi.
\newblock Robotic router formation in realistic communication environments.
\newblock \emph{IEEE Transactions on Robotics}, 28\penalty0 (4):\penalty0
  810--827, 2012.

\bibitem[Yan et~al.(2013)Yan, Jouandeau, and Cherif]{yan2013survey}
Zhi Yan, Nicolas Jouandeau, and Arab~Ali Cherif.
\newblock A survey and analysis of multi-robot coordination.
\newblock \emph{International Journal of Advanced Robotic Systems}, 10\penalty0
  (12):\penalty0 399, 2013.

\bibitem[Yang et~al.(2023)Yang, Lyu, and Luo]{yang2023minimally}
Yupeng Yang, Yiwei Lyu, and Wenhao Luo.
\newblock Minimally constrained multi-robot coordination with line-of-sight
  connectivity maintenance.
\newblock In \emph{2023 IEEE International Conference on Robotics and
  Automation (ICRA)}, pages 7684--7690. IEEE, 2023.

\bibitem[Yang et~al.(2024)Yang, Lyu, Zhang, Gao, and Luo]{yang2024integrating}
Yupeng Yang, Yiwei Lyu, Yanze Zhang, Ian Gao, and Wenhao Luo.
\newblock Integrating online learning and connectivity maintenance for
  communication-aware multi-robot coordination.
\newblock In \emph{2024 IEEE/RSJ International Conference on Intelligent Robots
  and Systems (IROS)}, pages 5770--5776. IEEE, 2024.

\bibitem[Zhang et~al.(2020)Zhang, Shu, Zhang, Ren, Zhou, and
  Chen]{zhang2020cellular}
Xin Zhang, Xiujun Shu, Bingwen Zhang, Jie Ren, Lizhou Zhou, and Xin Chen.
\newblock Cellular network radio propagation modeling with deep convolutional
  neural networks.
\newblock In \emph{Proceedings of the 26th ACM SIGKDD International Conference
  on knowledge discovery \& data mining}, pages 2378--2386, 2020.

\bibitem[Zhao et~al.(2023)Zhao, An, Pan, and Yang]{zhao2023nerf2}
Xiaopeng Zhao, Zhenlin An, Qingrui Pan, and Lei Yang.
\newblock Nerf2: Neural radio-frequency radiance fields.
\newblock In \emph{Proceedings of the 29th Annual International Conference on
  Mobile Computing and Networking}, pages 1--15, 2023.

\bibitem[Zhou et~al.(2023)Zhou, Xu, and Shen]{zhou2023racer}
Boyu Zhou, Hao Xu, and Shaojie Shen.
\newblock Racer: Rapid collaborative exploration with a decentralized multi-uav
  system.
\newblock \emph{IEEE Transactions on Robotics}, 39\penalty0 (3):\penalty0
  1816--1835, 2023.

\bibitem[Zhou et~al.(2021)Zhou, Zhu, Zhou, Xu, and Gao]{zhou2021ego}
Xin Zhou, Jiangchao Zhu, Hongyu Zhou, Chao Xu, and Fei Gao.
\newblock Ego-swarm: A fully autonomous and decentralized quadrotor swarm
  system in cluttered environments.
\newblock In \emph{2021 IEEE international conference on robotics and
  automation (ICRA)}, pages 4101--4107. IEEE, 2021.

\end{thebibliography}

\end{document}